\pdfoutput=1
\documentclass[11pt]{article}

\usepackage[final]{acl}

\usepackage{times}
\usepackage{latexsym}

\usepackage[utf8]{inputenc}

\usepackage{microtype}

\usepackage{inconsolata}

\usepackage{microtype}
\usepackage{graphicx}
\usepackage{subfigure}
\usepackage{booktabs} 
\usepackage{hyperref}

\usepackage{amsmath}
\usepackage{mathrsfs}
\usepackage{amssymb}
\usepackage{mathtools}
\usepackage{amsthm}
\usepackage{url}            
\usepackage{bm}
\usepackage{booktabs}       
\usepackage{amsfonts}       
\usepackage{nicefrac}       
\usepackage{microtype}      
\usepackage{adjustbox}
\usepackage{xcolor}         
\usepackage{blindtext}
\usepackage{enumitem}
\usepackage{listings}
\usepackage{pifont}
\usepackage{epsfig} 
\usepackage{times}
\usepackage{color}

\usepackage{multirow}
\usepackage{multicol}
\usepackage{makecell}
\usepackage{algorithm}
\usepackage{rotating}
\usepackage{subfigure}
\usepackage[T1]{fontenc}
\usepackage{algorithm}
\usepackage{algorithmic}
\usepackage{colortbl}
\usepackage{color, soul}
\usepackage{arydshln} 
\usepackage{tcolorbox}
\usepackage{ifsym}

\definecolor{babyblue}{rgb}{0.54, 0.81, 0.94}
\definecolor{bisque}{rgb}{1.0, 0.89, 0.77}
\definecolor{bshade}{rgb}{0.55,0.75,0.95}

\definecolor{mygray}{gray}{.6}
\definecolor{myblue}{RGB}{89,158,254}
\definecolor{mygreen1}{RGB}{81,150,111}
\definecolor{mygreen2}{RGB}{93,174,86}
\definecolor{myred}{RGB}{160,0,0}
\definecolor{myyellow}{RGB}{227,207,87}
\usepackage{caption}
\usepackage[capitalize,noabbrev]{cleveref}
\usepackage{threeparttable}
\usepackage{wrapfig}
\usepackage{fontawesome5} 
\theoremstyle{plain}

\theoremstyle{definition}

\theoremstyle{remark}

\definecolor{babyblue}{rgb}{0.54, 0.81, 0.94}
\definecolor{bisque}{rgb}{1.0, 0.89, 0.77}
\definecolor{bshade}{rgb}{0.55,0.75,0.95}

\definecolor{mygray}{gray}{.6}
\definecolor{myblue}{RGB}{89,158,254}
\definecolor{mygreen1}{RGB}{81,150,111}
\definecolor{mygreen2}{RGB}{93,174,86}
\definecolor{myred}{RGB}{160,0,0}
\definecolor{myyellow}{RGB}{227,207,87}
\usepackage{bbding}
\usepackage{enumitem}
\let\oldding\ding
\renewcommand{\ding}[2][1]{\scalebox{#1}{\oldding{#2}}}
\usepackage[textsize=tiny]{todonotes}

\newcommand{\methodname}{GSQ-Tuning}

\newcommand{\tablestyle}[2]{\setlength{\tabcolsep}{#1}\renewcommand{\arraystretch}{#2}\centering\footnotesize}
\newlength\savewidth\newcommand\shline{\noalign{\global\savewidth\arrayrulewidth
  \global\arrayrulewidth 1pt}\hline\noalign{\global\arrayrulewidth\savewidth}}
\definecolor{defaultcolor}{RGB}{89,158,254}
\newcommand{\default}[1]{\cellcolor{defaultcolor!15}{#1}}
\definecolor{ourscolor}{HTML}{E8E2F7}

\definecolor{baselinecolor}{gray}{.9}

\title{GSQ-Tuning: \underline{G}roup-\underline{S}hared Exponents Integer in \\ Fully \underline{Q}uantized Training for LLMs On-Device Fine-tuning}

\author{
 \textbf{Sifan Zhou\textsuperscript{1,2$\dagger$$\ddagger$}},
 \textbf{Shuo Wang\textsuperscript{1$\dagger$}},
 \textbf{Zhihang Yuan\textsuperscript{1$\dagger$}},
 \textbf{Mingjia Shi\textsuperscript{1}},
 \textbf{Yuzhang Shang\textsuperscript{3\textrm{\faEnvelope}}},
 \textbf{Dawei Yang\textsuperscript{1\textrm{\faEnvelope}}}
\\
\\
 \textsuperscript{1}Houmo AI,
 \textsuperscript{2}Southeast University,
 \textsuperscript{3}Illinois Institute of Technology
\\
}

\begin{document}
\maketitle

\def\thefootnote{$\dagger$}\footnotetext{Equal contribution}
\def\thefootnote{$\ddagger$}\footnotetext{Work done as an intern at Houmo AI}
\begin{abstract}
Large Language Models (LLMs) fine-tuning technologies have achieved remarkable results. However, traditional LLM fine-tuning approaches face significant challenges: they require large Floating Point (FP) computation, raising privacy concerns when handling sensitive data, and are impractical for resource-constrained edge devices. While Parameter-Efficient Fine-Tuning (PEFT) techniques reduce trainable parameters, their reliance on floating-point arithmetic creates fundamental incompatibilities with edge hardware.
In this work, we introduce a novel framework for on-device LLM fine-tuning that eliminates the need for floating-point operations in both inference and training, named \textbf{GSQ-Tuning}. At its core is the Group-Shared Exponents Integer format, which efficiently represents model parameters in integer format using shared exponents among parameter groups. When combined with LoRA-like adapters, this enables fully integer-based fine-tuning that is both memory and compute efficient. We demonstrate that our approach achieves accuracy comparable to BF16-based fine-tuning while significantly reducing 1.85$\times$ memory usage. Moreover, compared to FP8, our method can reduce $\sim$ 5$\times$ power consumption and $\sim$ 11$\times$ chip area with same performance, making large-scale model adaptation feasible on edge devices.
\end{abstract}
\section{Introduction}
Recent advances in Large Language Models (LLMs) have delivered impressive results in a variety of natural language tasks~\citep{touvron2023llama,touvron2023llama2,liu2023gpt}.
LLMs are typically trained in several stages, including large-scale pretraining followed by one or more fine-tuning phases~\cite{dubey2024llama,liu2024deepseek}. 
LLM fine-tuning approaches like supervised fine-tuning (SFT)~\citep{zhang2023instruction}, usually employ curated, high-quality corpora for refining the model with a standard language modeling~\cite{chiang2023vicuna}.

Despite their effectiveness, most LLM fine-tuning approaches require powerful cloud servers or GPUs equipped with large memory capacities. 
This poses two significant challenges in real-world settings: (1) uploading sensitive data to remote servers poses a fundamental privacy risk, and (2) in many practical scenarios, models must be deployed on resource-constrained edge devices—such as mobile processors or embedded AI accelerators—where memory and power budgets are tightly limited.  
\textit{Such constraints become critical in LLM's personalized applications, where data cannot be shared with the cloud and model updates must remain local to ensure privacy. Meeting these challenges thus necessitates on-device adaptation methods capable of preserving data privacy and functioning within the limited memory and compute budgets of edge hardware.}

Parameter-Efficient Fine-Tuning (PEFT)~\cite{han2024parameter} techniques such as LoRA~\cite{hu2021lora} and QLoRA~\cite{dettmers2023qlora} alleviate part of this burden by reducing trainable parameters to around 1\% of the original model. Unfortunately, they remain reliant on floating-point operations for both forward and backward passes, which clashes with edge-device constraints in three ways. 
First, during fine-tuning, weights, activation and gradients must be stored and updated in high-precision floating-point. It introduces additional overhead or even makes the LLM fine-tuning impractical on edge devices. Second, floating-point representations incur high memory overhead (e.g., FP16 doubles the memory cost compared to INT8; for a 7B-parameter model, this can surpass 20GB memory during fine-tuning process, presenting substantial challenges for mobile processors). Last but not least, commercial edge AI accelerators (e.g., Qualcomm Hexagon~\cite{QUALCOMM2024}) typically get peak throughput only on integers, leaving up to 84\% of compute units idle under FP16 training.

Therefore, eliminating floating-point arithmetic for fine-tuning would have a substantial impact on software, hardware, and application design for efficient on-device LLM adaptation~\citep{ARM2020,kim2021bert}. 
While previous studies on integer quantization~\citep{jacob2018quantization,kim2021bert,xiao2022smoothquant, yuan2023rptq} verify the feasibility of inference, they do not extend to gradient quantization, which is required for effective fine-tuning of LLMs at the edge.

In this paper, we propose a new framework for resource-efficient on-device LLM fine-tuning, termed \textbf{GSQ-Tuning}. Central to our method is the \emph{Group-Shared Exponents Integer} format, a novel quantization strategy that replaces floating-point with a specialized integer-based representation. 
We integrate this with parameter-efficient LoRA-like modules to enable fully on-device fine-tuning without incurring large memory and computation costs. 
We further examine this design through a Pareto frontier analysis, which demonstrates how various bits-rank settings impact the trade-off between fine-tuning memory costs and accuracy. Extensive experiments across models of varying scales, different fine-tuning datasets, and diverse tasks have demonstrated the effectiveness and generalizability.
We highlight our main contributions as follows:
\begin{itemize}
    \item  \textbf{Group-Shared Exponents Integer Quantization:} We introduce a quantization strategy that shares exponents among groups, thereby reducing the storage and computation overhead while still representing model parameters in integer format. Combined with LoRA-like adapters, our method supports fine-tuning under tight memory constraints. 
    \item \textbf{Integer Forward and Backward Computations:} By extending integer quantization pipelines beyond inference to include gradients, both forward and backward passes remain hardware-friendly and efficiently utilize integer-focused edge accelerators. 
    \item \textbf{Pareto Frontier for Quantization Bits and Low-rank:} We demonstrates how various bits-rank settings impact the trade-off between fine-tuning memory costs and accuracy through a Pareto frontier analysis. We empirically show that our approach achieves accuracy on par with FP16-based fine-tuning while dramatically lowering both 1.85$\times$ memory footprint. Furthermore, compared with FP8, at comparable performance levels, our method (GSE-INT5) reduces the power consumption of MAC unit by $\sim$ 5 $\times$ and decreases chip area by $\sim$ 11 $\times$ comparing to the origin.
\end{itemize}
\section{Method}
\label{sec: method}

In this section, we present \methodname, a fully quantized training method for on-device LLM fine-tuning. We begin by reviewing the fundamentals of LLM PEFT, highlighting the bottlenecks of implementing existing PEFT methods on device, and then review relevant neural network quantization literature (Sec.\ref{sec: background}). Building on these insights, we propose a new LLM fine-tuning framework—\emph{Group-Shared Exponents Integer in Fully Quantized Training}—for on-device scenarios. To enable this framework, we design two key components: (1) A \emph{Group-Shared Exponents Integer data format} to replace floating-point representations (Sec.\ref{sec: gse}). 
(2) A \emph{Fully Quantized Fine-tuning Framework} that leverages our new data format (Sec.\ref{sec: FQFT}). 
Finally, we explore the performance–efficiency trade-off in \methodname~via Pareto frontier analysis (Sec.\ref{sec:Pareto}) , providing practical guidance for its use.
\subsection{Preliminaries}
\label{sec: background}

\noindent\textbf{Low-rank Adaptation.} LoRA~\cite{hu2021lora} is a milestone method that injects trainable low-rank adapters into linear layers, allowing efficient fine-tuning while keeping the original parameters unchanged. Specifically, a LoRA linear layer is parameterized by a non-trainable weight matrix $\mathbf{W}\in\mathbb{R}^{oc\times ic}$, along with trainable components $\mathbf{A} \in\mathbb{R}^{r\times ic}$ and $\mathbf{B} \in\mathbb{R}^{oc\times r}$, where $r$ is a small integer. The input $\mathbf{X} \in \mathbb{R}^{b \times ic}$ and output $\mathbf{Y} \in \mathbb{R}^{b \times oc}$ correspond to a linear layer with $oc\times ic$ processing a batch of size $b$. Building on LoRA, QLoRA integrates it with 4-bit NormalFloat (NF4) quantization and Double Quantization (DQ)techniques, enabling the fine-tuning of a 65B parameter model on a single 48GB GPU with minimal performance loss. In this paper, due to the memory constraint of on-device PEFT, we adopt QLoRA to quantize the weights of LLMs. The formulation is:
\begin{equation*}
\small
\begin{aligned}
\vspace{-5mm}
    \label{eq:peft-qlora}
    \mathbf{Y = X \textit{DQ}(W^{NF4})^T + X A^T B^T}
\vspace{-5mm}
\end{aligned}
\end{equation*}
where we omit the transpose for similarity, $NF4$ means the 4 bit NormalFloat (NF) data type and and $DQ$ is \textit{Double Quantization} operation to map weights from NF4 to BF16 in QLoRA. The low-rank components $\mathbf{A}$ and $\mathbf{B}$ and input $\mathbf{X}$ remain in BF16 during fine-tuning process. However, QLoRA still does not suit on-device PEFT scenarios because the low-rank term $X A^T B^T$ remains in BF16, whereas most on-device hardware only supports integer operations. This limitation motivates our development of a new LoRA method that relies exclusively on integer operations.

\noindent\textbf{Quantization.} Quantization~\citep{Jacob_2018_CVPR} is a crucial technique that maps a floating-point number to a discrete interval using integer values. Given a floating-point (FP) tensor $\mathbf{x}$ (such as weights, activations or gradients),
the $b$-bits formulation is:
\begin{equation*}
\small
\begin{aligned}
\label{eq:quant}
    \mathbf{\mathbf{x}_{int}} = Q(\mathbf{x}) & = \mathrm{clamp}\left(\left\lfloor \frac{\mathbf{x}}{s} \right\rceil + z, 0, 2^b - 1\right) 
\vspace{-2mm}
\end{aligned}
\end{equation*}
where $s = \frac{\max(|\mathbf{x}|)}{2^{b-1}-1}$ is the scaling factor, $z$ denotes zero points, $\lfloor \cdot \rceil$ refers to the round-to-nearest, and the function $\mathrm{clamp}(\cdot)$ clips values outside the integer range $\left[q_{\min}, q_{\max} \right]$ respectively. $\left[q_{\min}, q_{\max} \right]$ is the quantization range determined by the bit width $b$, where $q_{\min}=-sz$ and $q_{\max}=-s(2^b-1-z)$. 
\noindent\textbf{Fully Quantized Training (FQT).} FQT involves quantizing all tensors—weights, activations, and gradients—needed for computation-intensive operations (like matrix multiplication) during both forward and backward propagation~\cite{wang2018training,yang2020training,zhu2020towards}. When the network's weights, activations, and gradients are each quantized to 8 bits, this is referred to as W8A8G8 quantization. Notably, FQT is different from Quantization-Aware Training (QAT), we also discuss the difference in Sec.~\ref{qat_fqt}.
\subsection{Group-Shared Exponents Integer}
\paragraph{Low-bitwidth Floating Point (FP).}
Floating-point numbers are a commonly used data representation in deep learning. For instance, FP16 represents each number using 16 bits. Recently, lower-bit floating-point representations, such as FP8, have been introduced into the training processes of deep learning models~\cite{micikevicius2022fp8,baalen2023fp8}. FP8 operates in two modes: the E4M3 and E5M2 formats. In these formats, E represents the number of exponent bits and M denotes that of mantissa bits.

\begin{figure}[t]
    \centering
    \includegraphics[width=\linewidth]{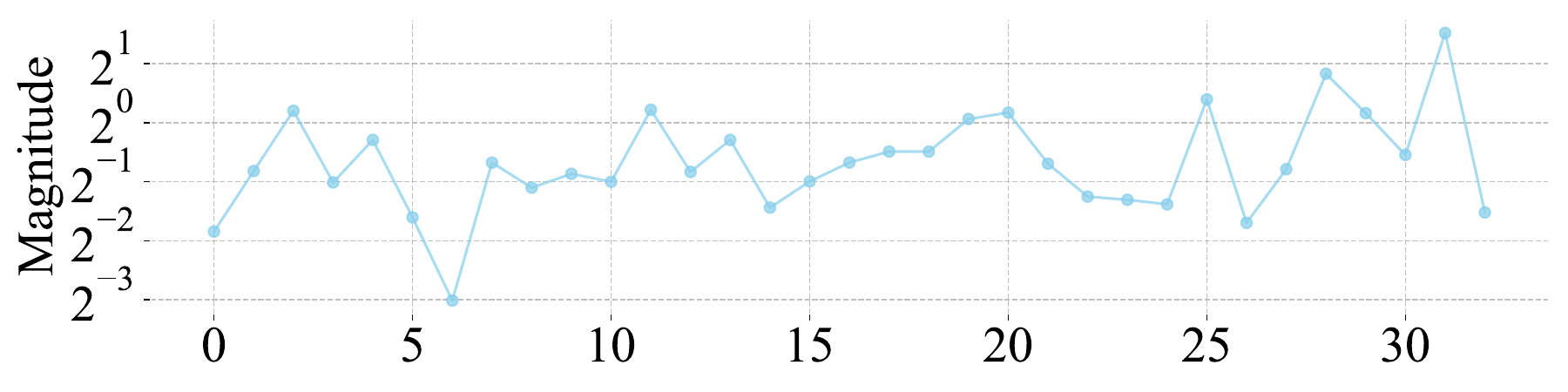}
    \vspace{-10pt}
    \includegraphics[width=\linewidth]{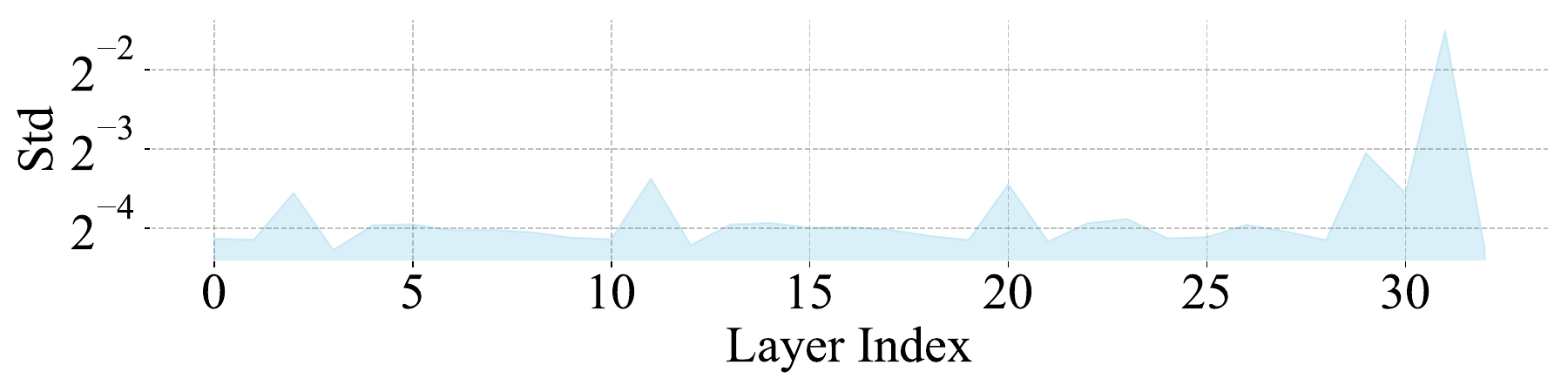}
    \caption{\textbf{In each layer, the weights' magnitudes are similar.} The standard deviations of weights across layers are less than $2^{-2}$ by 3-$\sigma$ (about probability 99.7\%). The weights are from Vicuna-7B-v1.5.}
    \label{fig:weights_std}
\end{figure}

Similar to quantization methods, low-bitwidth FP formats can effectively reduce memory storage requirements and decrease the hardware area and energy consumption of computational units. However, we observe that low-bitwidth FP may not be the optimal solution for LoRA fine-tuning in large-scale models, primarily due to the following 3 reasons: (1) Neural network tensors exhibit spatial locality, meaning that adjacent elements within a tensor tend to have similar magnitudes, leading to redundancy in the exponent bits of FP representations. As illustrated in Fig.~\ref{fig:weights_std}, the standard deviation of the values in the weight tensor is considerably lower than the magnitude of the values, indicating a small local variation; (2) The limited number of mantissa bits in low-bitwidth FP formats constrains precision, potentially impairing model performance. For instance, the E5M2 format, which has only two mantissa bits, is incapable of representing certain integers below ten, such as 5, 7, and 9; (3) FP computation demonstrates less efficiency in memory, chip area, and power consumption compared to GSE-INT computation, making it less suitable for resource-constrained environments. As shown in Table~\ref{tab:hardware_consumption}, FP formats incur considerably higher costs in power and chip area compared to integer-based computation, making them less suitable for edge environments.

Due to the inherent characteristics of FP representations and the requirement for relatively high precision in training, it is crucial to explore other data format to reduce both hardware area and energy consumption in resource-constrained cases.

\label{sec: gse}
\begin{figure}[t]
    \centering
    \includegraphics[width=\linewidth]{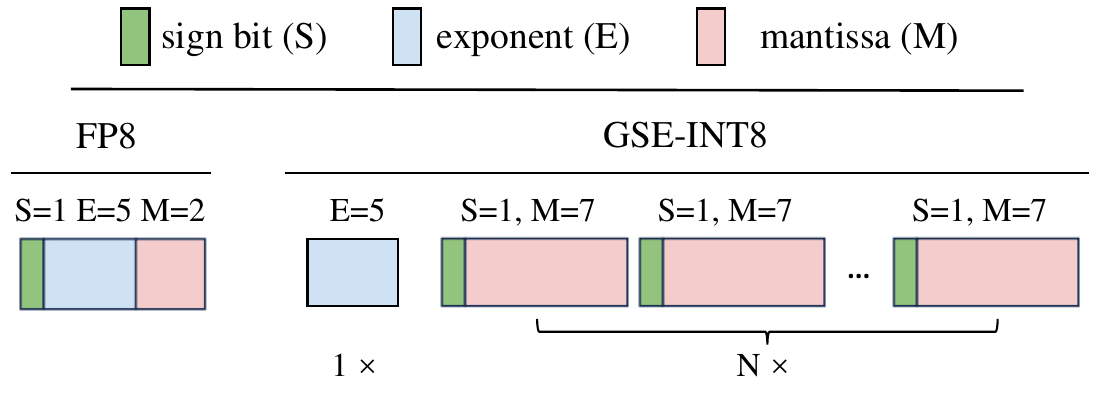}
    \caption{\textbf{The GSE format is memory efficient through group-shared exponent bits.} Comparison between FP8 and GSE-Int8.}
    \label{fig:comparison_format}
    \vspace{-4mm}
\end{figure}

\paragraph{Group-Shared Exponents Integer (GSE-INT).} 

Inspired with block FP~\cite{zhang2022fast}, we propose the Group-Shared Exponents Integer (GSE) format as an alternative to FP formats for matrix multiplication in both forward-propagation and back-propagation. 
This format is also used for storing activations required by back-propagation to reduce memory consumption. 
As illustrated in Fig.~\ref{fig:comparison_format}, GSE introduces the following key modifications compared to traditional floating-point formats:
(1) To leverage the locality of tensor values, we share the exponent across a group of N numbers. That is, all N numbers within the group use the same exponent.
(2) The number of bits used for the shared exponent is fixed at 5.
(3) The implicit leading 1 in floating-point representations is removed and replaced with a standard integer representation.
The numerical representation in GSE is:
\vspace{-2mm}
\begin{equation*}
\vspace{-2mm}
\begin{aligned}
    x = (-1)^{s} \cdot 2^{e} \cdot m
\end{aligned}
\end{equation*}
where $s$ is sign, $e$ is the exponent value (For simplicity, we omit the exponent bias), $m$ is the mantissa value.
The GSE format is memory efficient through sharing exponent bits. Memory for FP is $N(E+M+1)$ and memory for GSE is $N(M+1)+E$. As the group size N increases, the memory savings grow proportionally, while the overhead of the shared exponent is negligible.

\paragraph{Matrix Multiplication using GSE.}

Consider two vectors, \( \mathbf{A} \) and \( \mathbf{B} \), both represented using the GSE format and having a length of \( N \). The dot product of the two vectors can be computed as:
\vspace{-4mm}
\begin{equation*}
\small
\begin{aligned}
y &= 2^{e_A + e_B} \underbrace{\sum_{i=1}^N (-1)^{s_A \oplus s_B} m_{A,i} m_{B,i}}_{\mathclap{\text{standard integer multiply-accumulate}}},
\label{eq:GSE_dotprod}
\vspace{-2mm}
\end{aligned}
\end{equation*}
where \( m_{A,i} \) and \( m_{B,i} \) are the integer mantissas of the \( i \)-th elements of the vectors. The computation involves a standard integer multiply-accumulate (MAC) operation, followed by scaling with the combined exponent \( 2^{e_A + e_B} \).

The dot product operation can be extended to large-scale matrix multiplication. For two matrices \( \mathbf{X} \) and \( \mathbf{Y} \), we partition the data into groups of size \( N \). Specifically, rows of \( \mathbf{X} \) are grouped along their elements, with each group sharing a single exponent, and columns of \( \mathbf{Y} \) are grouped similarly. This grouping strategy simplifies hardware implementation and makes the GSE format a practical and efficient choice for large-scale matrix operations.

\paragraph{Transform from FP to GSE.} 

The transformation from FP representation to GSE format is efficient due to the design of GSE. First, within a group of \( N \) FP numbers, identify the largest exponent \( e_{\text{max}} \) among them. Then, double \( e_{\text{max}} \) to account for the shared exponent of the group. For each FP value in the group, its mantissa is adjusted by adding the implicit leading bit (if applicable) and then right-shifting the value based on the difference between its original exponent and \( e_{\text{max}} \). This process ensures that all values are aligned to the shared exponent, leading the storage and computation efficiency while preserving precision.


\begin{figure}[t]
  \centering
   \includegraphics[width=\linewidth]{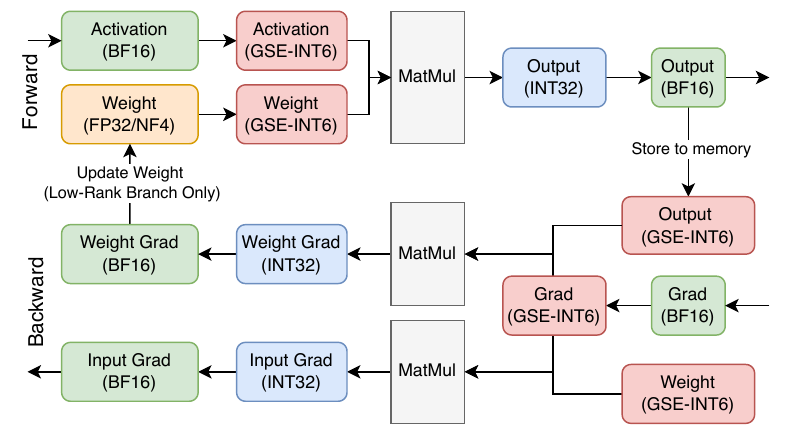}
   \caption{Dataflow of GSQ-Tuning. The weight is NF4 in full-rank branch and is FP32 in low-rank branch.}
   \label{fig:GSQ_tuning}
\vspace{-4mm}
\end{figure}

\subsection{Fully Quantized Fine-tuning}
\label{sec: FQFT}

As illustrated in Fig.~\ref{fig:GSQ_tuning}, our GSQ-Tuning framework introduces a hardware-efficient quantization pipeline. Compared to QLoRA, we fully quantize weights, activations, and gradients to low-bit integers. While QLoRA primarily focuses on 4-bit quantization of frozen base model weights (NF4) while keeping adapters in high precision (BF16), our approach achieves superior computational and memory efficiency. Building on the \textit{quantize-compute-dequantize} (QCD) paradigm for low-precision matrix multiplication (MM)~\citep{jetfire}, the QCD approach operates in three stages: (1) \textit{Quantization}: Convert high-precision inputs matrices (e.g., BF16) to low-precision (e.g., GSE-INT6) using a quantizer $Q(\cdot)$; (2) \textit{Computation in low-precision MM}: Perform low-precision MM to produce an intermediate output (e.g., GSE-INT6); and (3) \textit{Dequantization}: Convert output back to high-precision using a dequantizer $Q^{-1}(\cdot)$. 
\paragraph{Forward Propagation.} The forward propagation for a linear layer is calculated as follows:
\begin{equation*}
\small
\begin{aligned}
\label{forward}
    \mathbf{Y^{BF16}} & = \underbrace{Q^{-1}\left(Q(\mathbf{X^{BF16}}) Q\left(\textit{DQ}(\mathbf{W^{NF4}})\right)^T \right)}_{\text{frozen base model}} \\
    & \hfill + \underbrace{Q^{-1}\left(Q(\mathbf{X^{BF16}})Q(\mathbf{A^{BF16}})^TQ(\mathbf{B^{BF16}})^T \right)}_{\text{trainable adapter}}
\end{aligned}
\end{equation*}
\paragraph{Backward Propagation.}
Gradients are computed directly on quantized tensors using back propagation and chain rule:
\begin{equation*}
\small
\begin{aligned}
\vspace{-2mm}
\label{back_a}
\frac{\partial \mathcal{L}}{\partial \mathbf{A}} = Q^{-1}(Q(\mathbf{B})^TQ\left( \frac{\partial \mathcal{L}}{\partial \mathbf{Y}} \right)^T Q(\mathbf{X})) 
\end{aligned}
\end{equation*}

\vspace{-3mm}

\begin{equation*}
\small
\begin{aligned}
\vspace{-2mm}
\label{back_b}
\frac{\partial \mathcal{L}}{\partial \mathbf{B}} = Q^{-1}(Q\left(\frac{\partial \mathcal{L}}{\partial \mathbf{Y}}\right)^T Q(\mathbf{X})Q(\mathbf{A})^T) \\
\end{aligned}
\end{equation*}

\vspace{-3mm}

\begin{equation*}
\small
\begin{aligned}
\vspace{-2mm}
\label{back_x}
 \frac{\partial \mathcal{L}}{\partial \mathbf{X}} &= Q^{-1}(Q\left(\frac{\partial \mathcal{L}}{\partial \mathbf{Y}}\right)\left( Q(\mathbf{W}) + Q(\mathbf{B}) Q(\mathbf{A}) \right))
\end{aligned}
\end{equation*}

\subsection{Pareto Frontier for Quantization Bits and Low-rank.}
\label{sec:Pareto}

\paragraph{Co-optimization Principle for Model Bits and Rank.} The memory footprint and FLOPs during fine-tuning exhibit strong dependence on both quantization bit-width and LoRA rank $\mathcal{O}(b \cdot r)$ scaling. Excessive values in either dimension impose prohibitive computational burdens: (1) \textit{ Memory}:$\mathrm{Mem} \propto b \cdot r$ (adapter parameter storage); (2) \textit{ Compute}: $\mathrm{Flops} \propto r \cdot d^2$ (for hidden dimension $d$). This necessitates joint optimization of $(b,r)$ to guide the accuracy-efficiency trade-off space effectively. Pure bit-width reduction sacrifices model capacity, while unrestrained rank scaling inflates computation costs disproportionately.

The effectiveness of GSQ-Tuning hinges on how quantization bit-width interacts with the dimensions of low-rank adapters. To inform real-world deployments, we systematically analyze this interplay by constructing a Pareto frontier that illustrates the balance between model memory consumption during fine-tuning and accuracy across various bits-rank settings. We hope our findings not only highlight optimal configurations, but also offer practical guidelines for practitioners to tailor solutions to specific hardware constraints.

\begin{figure}[t]
  \centering
   \includegraphics[width=\linewidth]{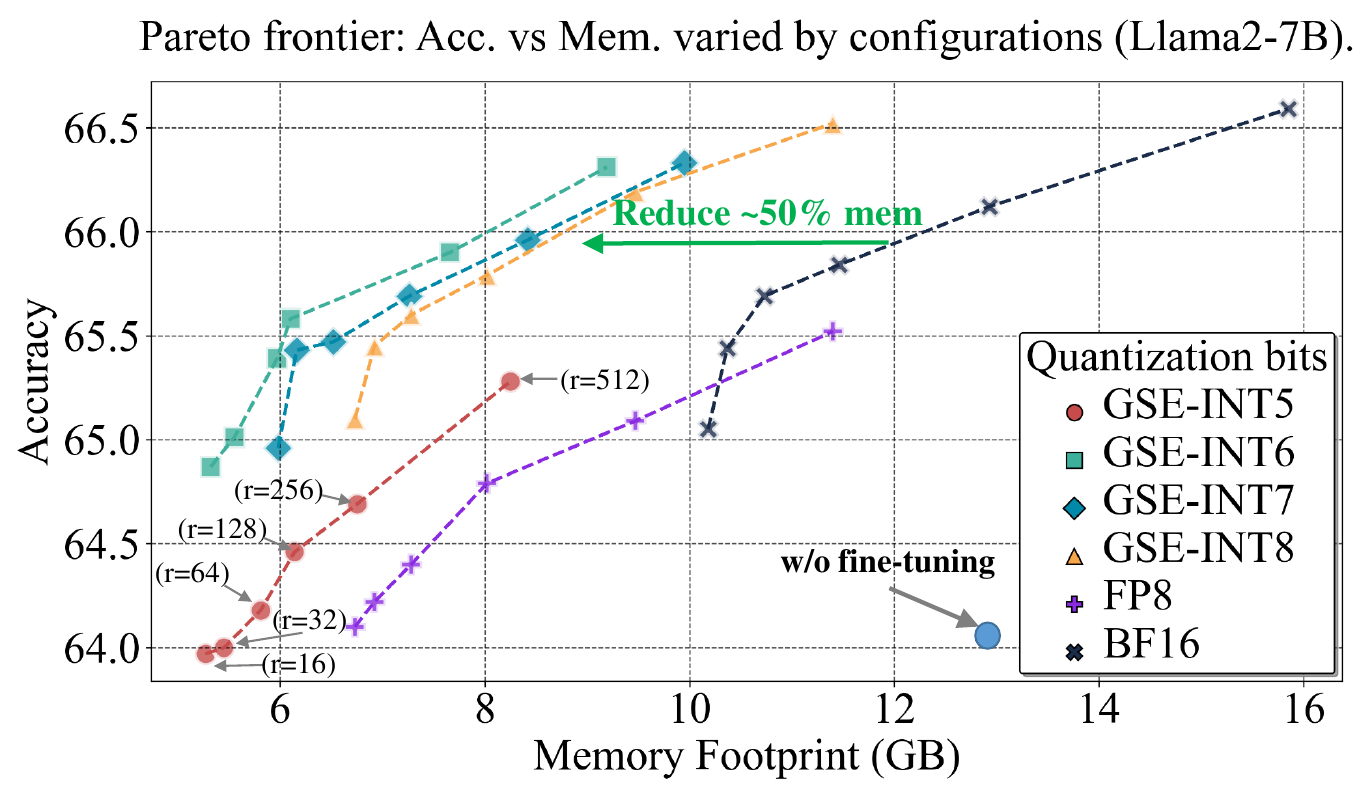}
   \caption{Pareto curve of accuracy-memory trade-offs. Compared to FP16, our GSQ-Tuning can reduce 1.85$\times$ memory usage while having the comparable accuracy. Detailed results are in Tab.\ref{tab:llama2-7b}}
    \label{fig:pareto}
    \vspace{-3mm}
\end{figure}

\paragraph{Pareto Frontier Analysis.} Based on our GSQ-Tuning, we construct a Pareto frontier by plotting model memory during fine-tuning against validation accuracy across different \textit{bits-rank} configuration. As shown in Fig.~\ref{fig:pareto}, the frontier reveals three distinct optimization regimes (1) \textit{High-Bit Low-Rank Regime (8-bit, r=64):} Reaches 65.60 Acc with suboptimal efficiency. 0.50 Acc gain from $r=16$ to $64$ indicates high-bit quantization inherently limits error magnitude, requiring less rank compensation. (2) \textit{Mid-Bit Balanced Regime (6-bit, r=128):} Delivers 65.58 Acc with moderate resources. 0.71 Acc gain from $r=16$ to $128$ shows diminishing returns beyond this point (only 0.32 Acc gain from $r=128$ to $512$). (3) \textit{Low-Bit High-Rank Regime (5-bit, r=512):}  Achieves 64.88 Acc with minimal memory footprint. 0.91 Acc gain from $r=16$ to $512$ demonstrates that aggressive rank scaling can compensate for severe quantization errors. Besides, Compared to FP8, GSQ-Tuning offers higher finetuning performance at similar or lower memory (Table~\ref{tab:copare_fp8_r32}) and reduces chip area ($\sim$11×) and power ($\sim$5×, Table~\ref{tab:hardware_consumption}). We also provide the Pareto frontier and detailed results for other models in appendix. Extra extensive results yield similar guidance.

\section{Experiments}
\label{sec:exp}

\begin{table*}[!t]
\renewcommand\arraystretch{1.0}
\centering
\caption{$0$-shot CSQA accuracy comparison with respect to different quantization bits in 64 rank setting. '4-8-8' means quantize weights, activation and gradients to 4-bit, 8-bit and 8-bit. GSQ-Tuning is built on Qlora, where all weights are quantized as NF4 firstly.Notably, the memory usage for LLaMA series model is for the model weights alone. Since these models are not fine-tuned, no fine-tuning phase (no gradient computation or updates) is involved. Denoted as 16-16-None for weights-activation-gradients.}
\vspace{-2mm}
\label{tab:compare_gsq}
\setlength{\tabcolsep}{1.2mm}
{\resizebox{0.98\textwidth}{!}{
\begin{tabular}{lccccccccccccc|c}
\noalign{\vspace{0.3em}}
\toprule
\noalign{\vspace{0.1em}}
\textbf{Method} & LLMs branch & low-rank branch  &\textbf{Avg.} & \textbf{ARC-c} & \textbf{ARC-e} & \textbf{BoolQ} & \textbf{HellaS.} & \textbf{OBQA} & \textbf{PIQA} & \textbf{SCIQ.} & \textbf{WinoG.} & \textbf{Mem.(G)} \\
\midrule
\noalign{\vspace{0.1em}}
 LLaMA2-7B         &  16-16-None & w/o  & \default{64.13} &46.25 &74.62 &77.68 &76.01 &44.20 &79.11 &46.11 &69.06&13.20\\
 \textit{w/ QLoRA} &  4-16-16 & 16-16-16 & \default{65.69} &47.14 &74.75 &79.50 &76.46 &45.50 &79.63 &50.26 &71.32&10.73\\
\noalign{\vspace{0.1em}}\hdashline[0.8pt/1pt]\noalign{\vspace{0.1em}}
\multirow{3}{*}{w/ GSQ-Tuning}  & 4-8-8 & 8-8-8  & \default{\textbf{65.60}} &48.12 &74.24 &79.72 &76.00 &45.80 &79.60 &49.69 &71.67& 7.28\\
& 4-6-6 & 6-6-6  & \default{\textbf{65.39}} &47.70 &74.58 &79.24 &76.05 &44.60 &79.60 &50.41 &70.96 & 5.97\\
& 4-5-5 & 5-5-5 & \default{\textbf{64.18}} &45.14 &72.69 &75.20 &75.27 &46.40 &79.65 &48.62 &70.48 & 5.81 \\
\hline
\noalign{\vspace{0.1em}}
 LLaMA2-13B         &  16-16-None & w/o  &\default{66.65} &48.81 &76.47 &82.45 &79.67 &44.80 &80.36 &48.31 &72.38 &25.7\\
 \textit{w/ QLoRA} &  4-16-16 & 16-16-16  & \default{67.61} &49.66 &77.23 &83.30 &78.95 &45.40 &80.74 &51.59 &73.24 &17.42\\
\noalign{\vspace{0.1em}}\hdashline[0.8pt/1pt]\noalign{\vspace{0.1em}}
 \multirow{3}{*}{w/ GSQ-Tuning}  & 4-8-8 & 8-8-8  & \default{\textbf{67.48}} &49.57 &77.40 &82.87 &78.88 &46.20 &80.90 &50.72 &73.32 & 11.99\\
& 4-6-6 & 6-6-6 & \default{\textbf{67.35}} &49.66 &77.27 &82.75 &78.66 &79.05 &80.90 &50.97 &73.16 & 10.89 \\
& 4-5-5 & 5-5-5 & \default{\textbf{66.97}} &49.91 &76.60 &81.87 &78.15 &46.20 &80.41 &49.54 &73.09 & 10.33\\
\hline
 LLaMA2-70B  & 16-16-None & w/o  & \default{70.68} &56.91 &80.05 &85.78 &83.59 &48.60 &82.48
&48.67 &79.40 & 137.42\\
 \textit{w/ QLoRA} & 4-16-16 & 16-16-16  &\default{72.22} &59.81 &82.20 &86.51 &83.89 &50.40 &83.13 &51.48 &80.35 &66.82 \\
\noalign{\vspace{0.1em}}\hdashline[0.8pt/1pt]\noalign{\vspace{0.1em}}
 \multirow{3}{*}{w/ GSQ-Tuning} & 4-8-8 & 8-8-8  &\default{\textbf{72.20}} &59.90 &82.32 &86.51 &83.90 &50.20 &83.08 &51.59 &80.11 & 52.17\\
 & 4-6-6 & 6-6-6 &\default{\textbf{72.10}} &59.39 &82.15 &86.51 &83.94 &50.00 &83.30 &50.92 &80.58 & 48.71\\
 & 4-5-5 & 5-5-5 & \default{\textbf{71.70}} &58.87 &81.48 &85.90 &83.91 &49.60 &82.81 &50.67 &80.43 & 46.98\\
\hline
 LLaMA3-3B  & 16-16-None & w/o  & \default{62.64} &45.90 &71.68 &73.00 &73.64 &43.20 &77.42 & 47.08 &69.22 &6.42 \\
 \textit{w/ QLoRA} & 4-16-16 & 16-16-16 & \default{64.11} &48.63 &74.07 &77.22 &73.12 &41.60 &78.51 &49.33 &70.40 & 6.78 \\
\noalign{\vspace{0.1em}}\hdashline[0.8pt/1pt]\noalign{\vspace{0.1em}}
 \multirow{3}{*}{w/ GSQ-Tuning} & 4-8-8 & 8-8-8  & \default{\textbf{64.02}} &47.30 &74.44 &77.26 &72.51 &42.40 &78.60 &49.80 &69.86& 3.93\\
 & 4-6-6 & 6-6-6 & \default{\textbf{63.71}} &47.07 &73.59 &76.64 &72.23 &41.80 &78.20 &48.76 & 71.36 & 3.47\\
 & 4-5-5 & 5-5-5 & \default{\textbf{62.74}} &48.04 &73.44 &73.36 &71.96 &40.40 &78.13 &47.65 & 68.98 & 3.24\\
\hline
 LLaMA3-8B         & 16-16-None & w/o  & \default{67.18} &53.50 &77.74 &81.13 &79.20 &45.00 &80.63 &47.03 &73.24 & 15.01\\
 \textit{w/ QLoRA} & 4-16-16 & 16-16-16 &\default{68.45} &55.63 &80.13 &83.67 &78.78 &44.80 &81.28 &50.41 &72.93 &11.64\\
\noalign{\vspace{0.1em}}\hdashline[0.8pt/1pt]\noalign{\vspace{0.1em}}
 \multirow{3}{*}{w/ GSQ-Tuning} & 4-8-8 & 8-8-8  &\default{\textbf{68.61}} &55.97 &80.22 &83.61 &78.68 &45.20 &81.50 &50.41 &73.32 & 7.63\\
 & 4-6-6 & 6-6-6 & \default{\textbf{68.22}} &55.55 &79.29 &83.67 &78.47 &44.80 &80.90 &50.05 &73.09 & 6.86\\
& 4-5-5 & 5-5-5 & \default{\textbf{66.69}} &54.10 &77.99 &81.65 &77.12 &43.80 &79.54 &47.90 &71.43 & 6.47\\
\hline
\end{tabular}}}
\vspace{-3mm}
\end{table*}

\noindent\textbf{Foundation Models and Evaluation Metrics.} We apply our method to the entire LLaMA family, including LLaMA-2 (7B/13B/70B)\citep{touvron2023llama2}, and LLaMA-3 (3B-8B). We evaluate the fine-tuning models on up to 9 zero-shot commonsense question-answering (CSQA) tasks using the \texttt{lm-evaluation-harness} (version 0.4.7)\citep{eval-harness}, including BoolQ\citep{clark2019boolq}, HellaSwag~\citep{zellers2019hellaswag}, LAMBADA (OpenAI)\citep{radford2019language}, OpenBookQA\citep{OpenBookQA2018}, PIQA~\citep{bisk2020piqa}, SIQA~\citep{sap2019socialiqa}, WinoGrande~\citep{sakaguchi2019winogrande}, ARC-Easy, and ARC-Challenge~\citep{clark2018think}. The fine-tuning dataset follows Alpaca~\citep{alpaca}, with 52K instruction data from text-davinci-003. Besides, we also report the memory cost (Mem.(G)) during the model fine-tuning process.

\noindent\textbf{Training Details.} We employ the whole fine-tuning process based on LLaMA-Factory~\citep{zheng2024llamafactory}. 
We implement \methodname{} in \texttt{PyTorch} using models from \texttt{Hugging Face}. We freeze the parameters of the linear modules and update a smaller (low-rank) set of parameters during the fine-tuning. The group size of our \methodname{} is $32$. We fine-tune models using the 8-bits AdamW optimizer~\citep{dettmers8} in bfloat16 precision. We choose the constant learning rate schedule and set the learning rate to be $1\times10^{-5}$ for all models. In all cases, we tune the hyper-parameters on the base BF16 tasks, and re-use the same values for low-precision training. We always perform single-epoch experiments using a linear learning rate warm-up of $100$ steps. The batch size and sequence length are fixed at $16$ and $2048$. The number of fine-tuning steps is 3.24K for Alpaca. 

\noindent\textbf{Hardware Synthesis.} We implemented the hardware in Verilog RTL and synthesized it using Synopsys Design Compiler with a 7nm technology library to estimate the process engine's area, latency, and power consumption~\citep{clark2016asap7}. The hardware operates at 1 GHz and has a capability of 50 TOPS. The memory subsystem is not considered in our settings and analyses.

\subsection{Overall Results}\label{subsec_overall_res}
\paragraph{GSQ-Tuning Results on LLaMA Family.} Here, we compare the fine-tuning performance across LLaMA family (3B$~$70B) against QLoRA. As shown in Tab.~\ref{tab:compare_gsq}, GSQ-Tuning achieves comparable or better zero-shot accuracy across different LLaMA model scales (7B-70B) under a fully low-bit quantization fine-tuing setting. With 8-bit quantization precision (W8A8G8), GSQ-Tuning matches or exceeds QLoRA's performance on 83\% of tasks, despite using 50\% fewer bits for activations and gradients. Even at aggressive 5-bit quantization (W5A5G5), the GSQ-Tuning maintains 98.6\% of QLoRA's average accuracy while reducing 1.85$\times$ memory footprint. These results confirm GSQ-Tuning's effectiveness for resource-constrained edge deployment. Besides, we present comprehensive results of our GSQ-Tuning across various LlaMA models in Sec.~\ref{sec:detailed_results}. This includes LlaMA2-7B (Tab.\ref{tab:llama2-7b}), LlaMA2-13B (Tab.\ref{tab:llama2-13b}), LlaMA2-70B (Tab.\ref{tab:llama2-70b}), LlaMA3-3B (Tab.\ref{tab:llama3-3b}), and LlaMA3-8B (Tab.\ref{tab:llama3-8b}). The findings consistently highlight the efficiency of GSQ-Tuning.

\begin{table*}[!t]
\renewcommand\arraystretch{1.0}
\centering
\caption{$0$-shot CSQA accuracy comparison with FP8 in different quantization bits in 32 rank setting.}
\label{tab:copare_fp8_r32}
\setlength{\tabcolsep}{1.2mm}
{\resizebox{0.98\textwidth}{!}{
\begin{tabular}{lccccccccccccc|c}
\toprule
\textbf{Method} & LLMs branch & low-rank branch &\textbf{Avg.} & \textbf{ARC-c} & \textbf{ARC-e} & \textbf{BoolQ} & \textbf{HellaS.} & \textbf{OBQA} & \textbf{PIQA} & \textbf{SCIQ.} & \textbf{WinoG.} & \textbf{Mem.(G)} \\
\hline
\noalign{\vspace{0.1em}}
 LLaMA2-7B          & 16-16-None & w/o  & \default{64.13} &46.25 &74.62 &77.68 &76.01 &44.20 &79.11 &46.11 &69.06 &13.2 \\
 \textit{w/ QLoRA}  & 4-16-16 & 16-16-16  &\default{65.24} &47.27 &75.04 &78.87 &76.11 &44.60 &79.76 &49.95 &70.32 &9.37 \\
\noalign{\vspace{0.1em}}\hdashline[0.8pt/1pt]\noalign{\vspace{0.1em}}
w/ FP8  & 4-8-8 & 8-8-8  & \default{64.09} &45.90 &73.27 &77.86 &75.83 &45.40 &78.45 &46.98 &69.06 &6.52 \\
\noalign{\vspace{0.1em}}\hdashline[0.8pt/1pt]\noalign{\vspace{0.1em}}
 \multirow{2}{*}{w/ GSQ-Tuning}  & 4-8-8 & 8-8-8  & \default{\textbf{65.45}} &48.12 &74.71 &78.38 &76.14 &46.00 &79.71 &49.64 &70.96 & 6.52\\
      & 4-5-5 & 5-5-5 &\default{\textbf{64.00}} &44.97 &73.32 &75.29 &74.95 &44.60 &79.27 &48.93 &70.24 & 5.45\\
\hline
 LLaMA3-8B  & 16-16-None & w/o  & \default{67.18} &53.50 &77.74 &81.13 &79.20 &45.00 &80.63 &47.03 &73.24 &15.01\\
 \textit{w/ QLoRA} &4-16-16 & 16-16-16 & \default{68.31} &55.55 &80.39 &83.36 &78.65 &44.60 &81.28 &50.05 &72.61 &11.02\\
\noalign{\vspace{0.1em}}\hdashline[0.8pt/1pt]\noalign{\vspace{0.1em}}
w/ FP8  & 4-8-8 & 8-8-8  &\default{66.62} &50.77 &76.43 &81.59 &78.17 &43.80 &80.20 &47.44 &74.59 &7.23 \\
\noalign{\vspace{0.1em}}\hdashline[0.8pt/1pt]\noalign{\vspace{0.1em}}
 \multirow{2}{*}{w/ GSQ-Tuning} & 4-8-8 & 8-8-8  & \default{\textbf{68.45}} &55.72 &80.22 &83.43 &78.60 &45.00 &81.18 &50.20 &73.32 & 7.23\\
& 4-5-5 & 5-5-5 & \default{\textbf{66.48}} &51.71 &77.69 &82.11 &76.91 &44.20 &79.43 &48.16 &71.67 & 6.07\\
\hline
\end{tabular}}}
\vspace{-2mm}
\end{table*}

\begin{table*}[t]
    \tablestyle{1pt}{1.3}
    \centering
    \caption{Cross-modal task evaluation on LLaVA-v1.5-7B: the {default} setting is QLoRA on 4-bits/64-rank without finetuning. Shared exponents shows robustness to LLM's finetuning, referring to the comparison with BF16.}
    \begin{tabular}{lcccccccccc}
    \noalign{\vspace{0.3em}}
    \toprule
    \noalign{\vspace{0.1em}}
    \multirow{2}{*}{\textbf{Settings}}
    &
    \multirow{2}{*}{LLMs branch} 
    &
    \multirow{2}{*}{low-rank branch}
    &
    \multicolumn{3}{c}{\textbf{POPE-random}}
    &
    \multicolumn{3}{c}{\textbf{POPE-adversarial}}
    &
    \multirow{2}{*}{\textbf{TextVQA}}
    &
    \multirow{2}{*}{\textbf{MMBench}}
    \\
    
     \cline{4-9}
    &
    &
    &
    accuracy
    &
    precision
    &
    F1-score
    & 
    accuracy
    &
    precision
    &
    F1-score
    & 
    \\
    \shline
    LLaVA-v1.5-7B 
    &
    16-16-None
    &
    w/o
    &
    {84.19}
    &
    {84.08}
    &
    {84.80}
    &
    {74.40}
    &
    {69.96}
    &
    {76.96}
    &
    {~~6.51}
    &
    {55.45}
    \\
    w/ QLoRA
    &4-16-16
    &
    16-16-16
    &
    87.47
    &
    92.51
    &
    86.68
    &
    83.87
    &
    85.52
    &
    83.48
    &
    47.68
    &
    67.08
    \\
    \hdashline[0.8pt/1pt]
    \multirow{2}{*}{w/ GSQ-Tuning}
    &
    4-8-8
    &
    4-8-8
    &
    {87.84}
    &
    {96.21}
    &
    {87.08}
    &
    \default{\textbf{84.03}}
    &
    \default{\textbf{87.40}}
    &
    \default{\textbf{83.28}}
    &
    {45.19}
    &
    {69.75}
    \\
    
    &
    4-6-6
    &
    4-6-6
    &
    \default{\textbf{88.08}}
    &
    \default{\textbf{96.08}}
    &
    \default{\textbf{87.39}}
    &
    83.43
    &
    85.80
    &
    82.87
    &
    \default{\textbf{49.13}}
    &
    \default{\textbf{70.19}}
    \\
    \hline
    \end{tabular}
    \label{tab:exp_mm}
    \vspace{-4mm}
\end{table*}
\begin{table*}[!t]
\renewcommand\arraystretch{1.0}
\centering
\small
\caption{$0$-shot CSQA accuracy on CS170K dataset in 64 rank setting.}
\vspace{-2mm}
\label{tab:cs170k}
\setlength{\tabcolsep}{1.2mm}
{\resizebox{0.98\textwidth}{!}{
\begin{tabular}{lccccccccccccc}
\toprule
\textbf{Method} & LLMs branch & low-rank branch &\textbf{Avg.} & \textbf{ARC-c} & \textbf{ARC-e} & \textbf{BoolQ} & \textbf{HellaS.} & \textbf{OBQA} & \textbf{PIQA} & \textbf{SCIQ.} & \textbf{WinoG.} \\
\hline
 LLaMA2-7B         & 16-16-None& w/o & \default{64.13} &46.25 &74.62 &77.68 &76.01 &44.20 &79.11 &46.11 &69.06\\
 w/ QLoRA  & 4-16-16& 16-16-16 &\cellcolor{myblue!15}67.78 & 51.79 & 78.86 & 81.28 & 75.77 & 46.00 & 79.98 & 53.89 & 75.37 \\
\noalign{\vspace{0.1em}}\hdashline[0.8pt/1pt]\noalign{\vspace{0.1em}}
 \multirow{2}{*}{w/ GSQ-Tuning}  & 4-8-8& 8-8-8  &\cellcolor{myblue!15}\textbf{67.73} & 51.79& 78.37 & 82.32 & 75.74 & 46.20 & 79.27 & 53.12 & 75.06 \\
      & 4-6-6& 6-6-6 & \cellcolor{myblue!15}\textbf{67.56}& 50.77 & 77.74 & 82.23 & 75.07 & 47.40 & 79.60 & 53.58 & 74.11 \\
\hline
\end{tabular}}}
\vspace{-4mm}
\end{table*}

\paragraph{Comparison with FP8.} Here, we compare the designed GSE data format with FP8 in fully quantized fine-tuning framework. As shown in Tab.~\ref{tab:copare_fp8_r32}, the results demonstrate that the designed GSE implemented in our GSQ-Tuning method achieves superior fine-tuning performance compared to FP8 while significantly reducing computation efficiency. Even under 5-bit settings, GSQ-Tuning maintains fine-tuning performance on par with FP8, further validating its effectiveness. Additionally, to mitigate the impact of rank variations, we report the fine-tuning results using 64-rank setting in Tab.~\ref{tab:copare_fp8_r64} of the appendix. Extensive experiments consistently support the advantages of our approach.

\begin{table}[h]
\renewcommand\arraystretch{1.0}
\centering
\caption{Fine-tuning time comparison with full-finetuning and low-rank adapter on alpaca-52K dataset in LLaMA2-7B model.}
\vspace{-2mm}
\label{tab:compare_time}
\setlength{\tabcolsep}{1.3mm}
{\resizebox{0.99\linewidth}{!}{
\begin{tabular}{lccccccccccccc|c}
\noalign{\vspace{0.3em}}
\toprule
\noalign{\vspace{0.1em}}
\textbf{Method} & LLMs branch & low-rank branch  &\textbf{Avg.} & \textbf{Ft-Time.(h)} & \textbf{Mem.(G)} \\
\midrule
\noalign{\vspace{0.1em}}
 LLaMA2-7B         &  16-16-None & w/o  & \default{64.13} &0.0&13.20\\
 \noalign{\vspace{0.1em}}\hdashline[0.8pt/1pt]\noalign{\vspace{0.1em}}
 \textit{w/ Full-finetuing} &  16-16-16 & 16-16-16 & \default{65.88} &2.4&58.56\\
 \textit{w/ QLoRA} &  4-16-16 & 16-16-16 & \default{65.69} &3.2&10.73\\
\noalign{\vspace{0.1em}}\hdashline[0.8pt/1pt]\noalign{\vspace{0.1em}}
\textit{w/ GSQ-Tuning}& 4-6-6 & 6-6-6  & \default{\textbf{65.39}} &3.4& 5.97\\
\hline
\end{tabular}}}
\vspace{-5mm}
\end{table}

\paragraph{Fine-tuning Time Comparison.} Here, we compare the fine-tuning time with QLoRA, full-finetune, and our proposed GSQ-Tuning on the LLaMA2-7B model. As shown in Tab.~\ref{tab:compare_time}, the results demonstrate that full fine-tuning achieves 65.88 accuracy in 2.4 hours but requires up to a typically infeasible 58.56GB memory ~10× our GSQ-Tuning’s 5.97GB and ~5× QLoRA’s 10.73GB. GSQ-Tuning (GSE-INT6) and QLoRA (BF16+NF4) both converge in 3.2~3.4 hours with matching accuracy (65.39\% vs 65.69\%), yet GSQ-Tuning uses nearly half QLoRA’s memory. It verifies that GSQ-Tuning matches QLoRA’s training time while offering superior memory efficiency, in resource-constrained edge scenarios where our primary focus. Although PEFT methods like ours inherently require longer convergence times, our primary focus lies in striking a balance between memory efficiency and computational feasibility.

\begin{table}[h]
\renewcommand\arraystretch{1.0}
\centering
\caption{Fine-tuning training memory (GB) cost with different fine-tuning settings in the LLaMA family.}
\vspace{-2mm}
\label{tab:compare_mem}
\setlength{\tabcolsep}{1.3mm}
{\resizebox{0.99\linewidth}{!}{
\begin{tabular}{lccccccccccccc|c}
\noalign{\vspace{0.3em}}
\toprule
\noalign{\vspace{0.1em}}
\textbf{Setting} & LLaMA2-7B &LLaMA2-13B &LLaMA2-70B &LLaMA3-3B &LLaMA3-8B\\
\midrule
\noalign{\vspace{0.1em}}
 \textit{Full-finetuing} &  58.56 & 112.80 &597.74 &28.12 &65.86\\
 \textit{w/ QLoRA} &  10.73& 17.42 &66.82 &6.78 &11.64\\
\noalign{\vspace{0.1em}}\hdashline[0.8pt/1pt]\noalign{\vspace{0.1em}}
\textit{w/ GSQ-Tuning}& \cellcolor{myblue!15}\textbf{5.97}&	\cellcolor{myblue!15}\textbf{10.89} &\cellcolor{myblue!15}\textbf{48.71}& \cellcolor{myblue!15}\textbf{3.47} & \cellcolor{myblue!15}\textbf{6.86}\\
\hline
\end{tabular}}}
\vspace{-5mm}
\end{table}

\paragraph{Fine-tuning Memory Comparison.} Here, as shown in Table~\ref{tab:compare_mem}, we reported the training memory cost for these baselines when training/full fine-tuning (batch size = 1, sequence length=2048), no activation checkpointing and memory offloading as shown in table R1. However, their huge training memory costs (>10x more than GSQ-Tuning) outside our edge-focused fine-tuning scope, where full finetuing/training is typically infeasible due to resource constraints.

\begin{table}[!t]
\renewcommand\arraystretch{1.0}
\centering
\caption{Comparison of hardware overhead between FP process engine and GSE-INT process engine (7nm).}
\label{tab:hardware_consumption}
\setlength{\tabcolsep}{1.2mm}
{\resizebox{0.75\linewidth}{!}{
\begin{tabular}{lcc}
\toprule
Format & Area ($\text{mm}^2$) &Power (W) \\
\hline
FP8 (E5M2)  &4.36 &2.53\\
FP8 (E4M3)  &5.06 &3.23\\
FP7 (E3M3)  &5.05 &2.75\\
FP6 (E3M2)  &3.40 &2.09\\
GSE-INT8  &\cellcolor{myblue!15}\textbf{0.85} &\cellcolor{myblue!15}\textbf{1.24}  \\
GSE-INT7  &\cellcolor{myblue!15}\textbf{0.61} &\cellcolor{myblue!15}\textbf{1.00} \\
GSE-INT6  &\cellcolor{myblue!15}\textbf{0.47} &\cellcolor{myblue!15}\textbf{0.76}\\
GSE-INT5  &\cellcolor{myblue!15}\textbf{0.39} &\cellcolor{myblue!15}\textbf{0.53}\\
\bottomrule
\end{tabular}}}
\vspace{-4mm}
\end{table}

\paragraph{Hardware Efficiency Analysis.} Table~\ref{tab:hardware_consumption} compares the chip area and power consumption of different formats through hardware synthesis. GSE-INT format demonstrates significant advantages over FP: (1) Area Efficiency: GSE-INT6 process engine requires only 0.47mm$^2$, 10.7$\times$ smaller than FP8(E4M3). This stems from simplified integer arithmetic logic and group-wise exponent sharing that eliminates complex alignment. (2) Power Superiority: At comparable bit-widths, GSE-INT6 consumes 0.76W (the 23.52\% of FP8's 3.23W).

\vspace{-3mm}
\subsection{Generalization Experiments}\label{subsec_generalization}
\paragraph{Generalization of GSQ-Tuning for Vision-Language Model (LLaVA).}
Model used is LLaVA-v1.5-7B~\citep{liu2024improved} with Vicuna-7B-v1.5~\citep{zheng2023judging} as language model and CLIP ViT-L-336px~\citep{radford2021learning} as vision tower, connected by a 2-layer MLP. Instruction dataset and other settings for finetuning follow the LLaVA official repository, LLaVA-Instruction~\citep{liu2024visual} and the improved one~\citep{liu2024improved}.
Tab.~\ref{tab:exp_mm} shows performance drop of the vanilla quantization of 4-bits/64-rank QLoRA, especially, referring to the TextVQA evaluation. Finetuning with GSE shows comparable performance compared to that with BF16. BF16 is of E8M7 while GSE is of E5M7, demonstrating the redundancy of the dynamic range \textit{w.r.t.} exponents is at least 3-bits much. Moreover, the memory cost of GSE is about a half of BF16.

\paragraph{Generalization of GSQ-Tuning on Other Fine-tune Dataset.} Here, we also select Commonsense170K (CS170K)~\citep{hu2023llmadapters} to evaluate the generalization ability of GSQ-Tuning across different fine-tuing dataset. CS170K is a dataset constructed from the training sets of BoolQ, PIQA, SIQA, HellaSwag, WinoGrande, ARC-e, ARC-c, and OBQA with pre-defined templates, comprising 170K commonsense reasoning samples. As shown in Tab.~\ref{tab:cs170k}, on larger fine-tuning datasets, our GSQ-Tuning also demonstrates comparable or even superior accuracy compared to QLoRA, while being more computationally efficient.

\vspace{-3mm}
\subsection{Ablation Study.}\label{ablation}
\begin{table}[!t]
\renewcommand\arraystretch{1.0}
\centering
\caption{The effect of the number of shared group on fine-tuning performance in 64 rank setting.}
\label{tab:ablation_group}
\setlength{\tabcolsep}{1.2mm}
{\resizebox{0.99\linewidth}{!}{
\begin{tabular}{lccccc}
\noalign{\vspace{0.3em}}
\toprule
\noalign{\vspace{0.1em}}
\textbf{Method} & LLMs branch & low-rank branch &\textbf{Group} &\textbf{Avg.} & \textbf{Mem. (G)} \\
\midrule
 LLaMA2-7B      &16-16-None & w/o   & - & 64.13 &13.2\\
\noalign{\vspace{0.1em}}\hdashline[0.8pt/1pt]\noalign{\vspace{0.1em}}
\noalign{\vspace{0.1em}}
\multirow{3}{*}{w/ GSQ-Tuning}  &\multirow{3}{*}{5-6-6} &\multirow{3}{*}{6-6-6}  & 32& \cellcolor{myblue!15}\textbf{65.39} &6.17\\
 &   & & 64&\cellcolor{myblue!15}64.72 &6.32 \\
 &   & & 128&\cellcolor{myblue!15}64.27 &6.56\\
\bottomrule
\end{tabular}}}
\vspace{-2mm}
\end{table}
\paragraph{Group Size Analysis.} As shown in Tab.~\ref{tab:ablation_group}, our GSQ-Tuning with 32 groups achieves optimal accuracy-efficiency balance in 6-bit configurations (W6A6G6). The 32-group setting yields significantly higher average accuracy (65.39) compared to 64-group (64.72) and 128-group (64.27) variants, while maintaining comparable memory efficiency (70.96 vs 69.22/69.53). This sweet spot emerges from the tension between quantization bit width and hardware deployment - smaller groups better capture value distributions but increase computation overhead, while larger groups sacrifice adaptation granularity. We therefore adopt group=32 as the default configuration.

\begin{table}[h]
\renewcommand\arraystretch{1.0}
\centering
\caption{The ablation of the exponent bit width (E) and the number of shared group (N) on fine-tuning performance in 64 rank setting.}
\label{tab:ablation_bits}
\setlength{\tabcolsep}{1.2mm}
{\resizebox{0.99\linewidth}{!}{
\begin{tabular}{lccccc}
\noalign{\vspace{0.3em}}
\toprule
\noalign{\vspace{0.1em}}
\textbf{Method} & LLMs branch & low-rank branch &\textbf{Group} &\textbf{Avg.} & \textbf{Mem. (G)} \\
\midrule
LLaMA2-7B      &16-16-None & w/o   & - & 64.13 &13.2\\
\noalign{\vspace{0.1em}}\hdashline[0.8pt/1pt]\noalign{\vspace{0.1em}}
\noalign{\vspace{0.1em}}
\multirow{6}{*}{w/ GSQ-Tuning}  &\multirow{2}{*}{4-6-6} &\multirow{2}{*}{6-6-6}  & 32& \cellcolor{myblue!15}62.85 &6.29\\
 &   & & 64&\cellcolor{myblue!15}62.66 &6.16 \\
 &\multirow{2}{*}{5-6-6} &\multirow{2}{*}{6-6-6}  & 32& \cellcolor{myblue!15}\textbf{65.39} &6.32\\
 &   & & 64&\cellcolor{myblue!15}64.72 &6.17 \\
 &\multirow{2}{*}{6-6-6} &\multirow{2}{*}{6-6-6}  & 32& \cellcolor{myblue!15}65.41 &6.35\\
 &   & & 64&\cellcolor{myblue!15}64.99 &6.18 \\
\bottomrule
\end{tabular}}}
\vspace{-6mm}
\end{table}
\paragraph{Exponenet Bits Analysis.} As shown in Tab.~\ref{tab:ablation_bits}, experiments demonstrate that fixing exponent bits E=5 achieves the optimal trade-off between accuracy and memory overhead. Adopting a smaller exponent (for example, E = 4) leads to significant performance degradation, while increasing E=6 yields marginal performance gains. This ablation study validates the effectiveness of our fixed shared exponent design E = 5.

\begin{table}[!t]
\renewcommand\arraystretch{1.0}
\centering
\caption{The effect of the number of low rank on fine-tuning performance in group=32 setting.}
\vspace{-2mm}
\label{tab:ablation_rank}
\setlength{\tabcolsep}{1.2mm}
{\resizebox{0.85\linewidth}{!}{
\begin{tabular}{lcccc}
\noalign{\vspace{0.3em}}
\toprule
\noalign{\vspace{0.1em}}
\textbf{Method} & LLMs branch &low-rank branch &\textbf{Rank} &\textbf{Avg.}\\
\hline
 LLaMA2-7B         & 16-16-None & w/o&- & 64.13 \\
\noalign{\vspace{0.1em}}\hdashline[0.8pt/1pt]\noalign{\vspace{0.1em}}
\noalign{\vspace{0.1em}}
\multirow{6}{*}{w/ GSQ-Tuning}  & \multirow{6}{*}{5-6-6} & \multirow{6}{*}{6-6-6}  & 16&\cellcolor{myblue!15}64.87 \\
 & & & 32&\cellcolor{myblue!15}65.01  \\
 & &  & 64& \cellcolor{myblue!15}65.39 \\
 & &  & 128& \cellcolor{myblue!15}65.58 \\
  & &  & 256&\cellcolor{myblue!15}65.90 \\
   & &  & 512&\cellcolor{myblue!15}\textbf{66.31}\\
\bottomrule
\end{tabular}}}
\vspace{-4mm}
\end{table}
\paragraph{Low Rank Size Analysis.} As shown in Tab.~\ref{tab:ablation_rank}, the results demonstrates the accuracy-efficiency trade-off across different LoRA ranks in 6-bit configurations (W6A6G6). While larger ranks consistently improve performance (64.87 $\rightarrow$ 66.31 average accuracy for rank 16 $\rightarrow$ 512), the marginal gains diminish significantly beyond rank 64 (+0.38 from 64→128 vs +0.73 from 16 $\rightarrow$ 64). This aligns with the spectral properties of weight matrices, where most task-relevant information is captured by the dominant singular vectors. We observe that rank=64 provides an optimal balance, achieving 98.6\% of the maximum accuracy while using 50\% less memory than rank=512 configurations.

\section{Related Work}
\label{related_works}
\vspace{-2mm}

\paragraph{Parameter-Efficient Fine-Tuning (PEFT).} PEFT reduces memory and computational costs by introducing a small set of trainable parameters while keeping the pretrained model frozen. Approaches including soft prompt tuning \citep{wang2023multitask}, partial fine-tuning \citep{fu2023effectiveness}, and low-rank adaptation \citep{hu2021lora}. Among these, LoRA stands out as a seminal work, injecting trainable low-rank matrices into linear layers to enable efficient fine-tuning without modifying the base model weights. QLoRA extends this with 4-bit NF4 quantization and Double Quantization, supporting 65B model fine-tuning on a single 48GB GPU with minimal performance degradation. Despite recent studies further extending LoRA from the perspective of quantization-aware fine-tuning \citep{xu2023qa, liloftq} to improve efficiency, stability, and performance \citep{hu2023llmadapters,liu2024dora, zhao2024galore,hayou2024lora+,meng2024pissa}, these methods still maintain compute-intensive forward and backward propagations at high bit-widths during fine-tuning. This results in a substantial computational burden when targeting edge AI accelerators.

\paragraph{Quantization.} Much excellent works ~\cite{yuan2023rptq,shang2023pbllm,yue2024wkvquant, frantar2022gptq, xiao2022smoothquant,hu2024illm,hu2025ostquant,hu2025moequant} use the quantization techniques to accelerate the inference of LLMs. For instance, GPTQ~\cite{frantar2022gptq} quantizes weights to 3-4 bit with slight accuracy drop based on approximate second-order information. AWQ~\cite{lin2024awq}, SmoothQuant~\cite{xiao2022smoothquant}, and OmniQuant~\cite{shaoomniquant} explore the scheme of smoothing by detecting the importance of different activation channels. Recent works (e.g., Quarot~\cite{ashkboos2024quarot}, SpinQuant~\cite{liu2024spinquant}, OSTQuant~\cite{hu2025ostquant}) further suppress outliers by utilizing computation-invariant rotation transformation. However, above methods mainly focus on the post-training optimization while overlooking the overhead during training. Interestingly, several studies~\cite{banner2018scalable,drumond2018training,adelman2018faster,wu2018training,langroudi2019deep,langroudi2019cheetah,yang2020training,zhu2020towards,xi2023training} have made much efforts to improve the efficiency and optimization of the training process, notably through \textbf{Fully Quantized Training (FQT)}. FQT involves quantizing all tensors—weights, activations, and gradients—needed for computation-intensive operations (like matrix multiplication) during both forward and backward propagation~\cite{wang2018training,yang2020training,zhu2020towards}. Particularly during the backward propagation, quantizing gradients remains challenging due to the wide dynamic range. Recent works have focused on integer-based quantization for smaller-scale models; for instance, SwitchBack~\citep{switchback} quantizes partial matrix multiplications with INT8, but it is limited to vision models with up to 1B parameters. Jetfire~\citep{jetfire} proposes a 2D block-wise quantization approach that maintains accuracy and achieves significant memory savings (1.4-1.5x) when training in INT8. However, these methods have yet to scale to LLMs. Additionally, some efforts have attempted to alleviate the training burden using low-bit floating-point representations; for example, LM-FP8~\citep{peng2023fp8} trains LLMs from scratch using FP8, achieving performance comparable to BF16. Nonetheless, above methods have not yet been explored to for low-bit integer fine-tuning tasks for LLMs. Notably, FQT is different from Quantization-Aware Training (QAT), we also discuss the difference in Sec.~\ref{qat_fqt}.

\section{Conclusion} 
\label{sec:conclusion}
\vspace{-2mm}
In this paper, we propose GSQ-Tuning, a resource-efficient framework that addresses the critical challenges of floating-point dependency, privacy risks, and hardware incompatibility in on-device LLM fine-tuning. By integrating Group-Shared Exponents Integer (GSE) quantization with parameter-efficient adaptation, our method achieves three key advancements:(1) Full Integer Pipeline: Eliminates floating-point operations across both forward and backward passes, reducing 1.85$\times$ memory footprint compared to FP16 while maintaining comparable accuracy. (2) Hardware-Optimized Design: The GSE format reduces metadata overhead via group-wise exponent sharing, enabling 5-8bit integer representations. Combined with LoRA-like adapters, this achieves 5 $\times$ lower power consumption and 11$\times$ smaller chip area compared to FP8 at equivalent accuracy levels. (3) Practical Deployment Guidance: A Pareto frontier analysis guides optimal bit-rank configurations for diverse edge constraints. These innovations establish GSQ-Tuning as a foundational step toward democratizing LLM adaptation for resource-constrained environments. This breakthrough makes private, on-device LLM adaptation practical for sensitive applications. Future work will explore sub-4bit quantization to further push the boundaries of edge AI.

\section{Limitations and Future Work}

\paragraph{Limitations} While our GSQ-Tuning significantly advances on-device LLM adaptation through integer-focused optimization and parameter-efficient quantization, two key limitations warrant discussion:

\paragraph{Non-linear Operator Precision.} Our current implementation maintains non-linear operations (e.g., LayerNorm, Softmax) in 16-bit to preserve numerical stability. This introduces partial precision conversion overhead during computation. However, non-linear operations do not contain additional learnable parameters and thus do not consume memory. Moreover, these non-linear operations are generally computation-light, making their computational burden negligible. Future work could explore fully integer implementations for non-linear layers. \textbf{Bit-Width Range Constraints.}
The current framework operates effectively in 5-8bit configurations but didn't present the performance at extreme low bit ($\leq$ 4bit) precision. This stems from gradient direction distortion under extreme quantization—a challenge requiring new error compensation mechanisms. We plan to investigate two directions: (1) 4bit stochastic rounding with gradient-aware scaling, and (2) mixed-precision adapters allocating higher bits to critical gradient dimensions.

\paragraph{Future Work} Furthermore, future work could explore (1) full integer fine-tuning, (2) extreme low-bit quantized fine-tuning and (3) co-design with emerging integer-optimized AI accelerators.
\bibliography{custom}

\appendix

\section{Appendix}
\label{sec:appendix}
\subsection{Differences with Quantization-aware training (QAT): }
\label{qat_fqt}
Quantization-aware training (QAT)~\cite{choi2018pact,Zhang_2018_ECCV,zhou2017incremental,jacob2018quantization,dong2019hawq,dong2019hawqv2,shen2019q,zafrir2019q8bert,shen2020QBERT,tang2022mkq,zhang2020ternarybert,bai2020binarybert,foret2020sharpness,wang2022squat} is an \emph{inference acceleration} technique which trains networks with quantizers inserted in the forward propagation graph, so the trained network can perform efficiently during inference. 
QAT can compress activation/weights to extremely low precision (e.g. 1-2 bits). 
It is tempting to think that directly applying a quantizer for QAT to FQT can lead to similar low activation/weights bit-width. However, even only quantizing the forward propagation for FQT is much more challenging than QAT because:  \raisebox{-0.5pt}{\ding[1.1]{182\relax}} QAT requires a converged full-precision model as initialization~\cite{esser2019learned} and/or as a teacher model for knowledge distillation~\cite{bai2020binarybert}; \raisebox{-0.5pt}{\ding[1.1]{184\relax}} QAT may approximate the discrete quantizer with continuous functions during training~\cite{gong2019differentiable}, which cannot be implemented with integer arithmetic. Due to these challenges, it is still an open problem to do FQT with low-bit activations/weights. 

\begin{table*}[h]
\renewcommand\arraystretch{1.0}
\centering
\caption{$0$-shot commonsense QA accuracy (\%) across different bits and rank on llama2-7B.}
\label{tab:llama2-7b}
\setlength{\tabcolsep}{1.2mm}
{\resizebox{0.98\textwidth}{!}{
\begin{tabular}{lcccccccccccccc|c}
\noalign{\vspace{0.3em}}
\toprule
\noalign{\vspace{0.1em}}
\textbf{Method} & \textbf{rank}& LLMs branch & low-rank branch &\textbf{Avg.} & \textbf{ARC-c} & \textbf{ARC-e} & \textbf{BoolQ} & \textbf{HellaS.} & \textbf{OBQA} & \textbf{PIQA} & \textbf{SCIQ.} & \textbf{WinoG.} & \textbf{Mem. (G)} \\
\midrule
\noalign{\vspace{0.1em}}
 LLaMA2-7B    & &16-16-None &w/o & 64.13 &46.25 &74.62 &77.68 &76.01 &44.20 &79.11 &46.11 &69.06 &13.2\\
 \noalign{\vspace{0.1em}}\hdashline[0.8pt/1pt]\noalign{\vspace{0.1em}}
 \textit{w/ QLoRA} & 16 & 4-16-16 &16-16-16 & 65.05 &47.53 &75.17 &78.59 &76.09 &44.00 &79.54 &49.44 &70.09&10.18\\
\noalign{\vspace{0.1em}}\hdashline[0.8pt/1pt]\noalign{\vspace{0.1em}}
\multirow{4}{*}{w/ GSQ-Tuning}  &\multirow{4}{*}{16} & 4-8-8 &8-8-8  & \default{\textbf{65.10}}& 47.53 &74.71 &78.35 &75.99 &45.00 &79.65 &49.28 &70.32&6.73\\
& & 4-7-7 &7-7-7  & \default{\textbf{64.96}} &47.18 &75.21 &78.10 &75.98 &44.80 &79.27 &49.95 &69.38 &5.98\\
& & 4-6-6 &6-6-6  & \default{\textbf{64.87}} &46.84 &73.78 &78.07 &75.88 &45.80 &79.22 &49.39 &70.01 &5.32\\
  & & 4-5-5 &5-5-5 & \default{63.97} &46.76 &72.64 &75.78 &74.95 &45.20 &79.05 &48.62 &68.75 &5.27\\
\midrule
\noalign{\vspace{0.1em}}
 \textit{w/ QLoRA} & 32 & 4-16-16 &16-16-16 & 65.44 &47.27 &75.04 &78.87 &76.11 &44.60 &79.76 &49.95 &70.32 &10.37\\
\noalign{\vspace{0.1em}}\hdashline[0.8pt/1pt]\noalign{\vspace{0.1em}}
\multirow{4}{*}{w/ GSQ-Tuning}  &\multirow{4}{*}{32} & 4-8-8 &8-8-8  & \default{\textbf{65.45}} &48.12 &74.71 &78.38 &76.14 &46.00 &79.71 &49.64 &70.96 &6.92\\
& & 4-7-7 &7-7-7  & \default{\textbf{65.43}} &47.35 &74.20 &78.99 &75.84 &46.00 &79.92 &49.59 &71.59 &6.16\\
& & 4-6-6 &6-6-5  & \default{\textbf{65.01}} &47.44 &74.62 &78.65 &76.03 &44.00 &79.60 &50.05 &69.69 &5.55\\
  & &4-5-5 &5-5-5 & \default{64.00} &44.97 &73.32 &75.29 &74.95 &44.60 &79.27 &48.93 &70.24&5.45\\
\midrule
 \textit{w/ QLoRA} & 64 & 4-16-16 &16-16-16 & 65.69 &47.14 &74.75 &79.50 &76.46 &45.50 &79.63 &50.26 &71.32 &10.73\\
\noalign{\vspace{0.1em}}\hdashline[0.8pt/1pt]\noalign{\vspace{0.1em}}
\multirow{4}{*}{w/ GSQ-Tuning}  &\multirow{4}{*}{64} & 4-8-8 &8-8-8  & \default{\textbf{65.60}} &48.12 &74.24 &79.72 &76.00 &45.80 &79.60 &49.69 &71.67 &7.28\\
& & 4-7-7 &7-7-7  & \default{\textbf{65.47}} &47.78 &74.71 &79.51 &76.09 &45.80 &79.60 &49.80 &70.48 &6.52\\
& & 4-6-6 &6-6-6  & \default{\textbf{65.39}} &47.70 &74.58 &79.24 &76.05 &44.60 &79.60 &50.41 &70.96 &5.97\\
  & & 4-5-5 &5-5-5 & \default{64.18} &45.14 &72.69 &75.20 &75.27 &46.40 &79.65 &48.62 &70.48 &5.81\\
\midrule
 \textit{w/ QLoRA} & 128 & 4-16-16 &16-16-16& 65.84 &48.24 &74.91 &79.78 &76.27 &45.52 &79.77 &50.48 &71.79 &11.46\\
\noalign{\vspace{0.1em}}\hdashline[0.8pt/1pt]\noalign{\vspace{0.1em}}
\multirow{4}{*}{w/ GSQ-Tuning}  &\multirow{4}{*}{128} & 4-8-8 &8-8-8  & \default{\textbf{65.79}} &48.12 &74.83 &80.28 &75.96 &45.80 &79.54 &50.61 &71.19 &8.02\\
& & 4-7-7 &7-7-7  & \default{\textbf{65.69}} &48.04 &74.87 &79.79 &76.08 &45.00 &79.49 &50.61 &71.67 &7.26\\
& & 4-6-6 &6-6-6  & \default{\textbf{65.58}} &47.87 &74.54 &80.09 &76.05 &45.40 &79.38 &50.10 &71.27 &6.10\\
  & &4-5-5 &5-5-5 & \default{64.46} &46.50 &72.77 &75.99 &75.31 &46.60 &79.00 &48.98 &70.56 &6.14\\
\midrule
 \textit{w/ QLoRA} & 256 & 4-16-16 &16-16-16 & 66.12 &48.33 &75.00 &80.94 &76.37 &45.61 &79.97 &51.13 &71.64 &12.93\\
\noalign{\vspace{0.1em}}\hdashline[0.8pt/1pt]\noalign{\vspace{0.1em}}
\multirow{4}{*}{w/ GSQ-Tuning}  &\multirow{4}{*}{256} & 4-8-8 &8-8-8  & \default{\textbf{66.19}} &48.55 &75.13 &80.76 &76.14 &47.00 &79.38 &50.72 &71.82 &9.47\\
& & 4-7-7 &7-7-7  & \default{\textbf{65.96}} &48.46 &75.08 &80.43 &76.04 &45.60 &79.76 &50.72 &71.59 &8.42\\
& & 4-6-6 &6-6-6  & \default{\textbf{65.90}} &48.38 &74.16 &79.94 &75.81 &46.80 &79.43 &50.87 &71.82 &7.66\\
  & & 4-5-5 &5-5-5 & \default{64.59} &46.33 &72.60 &76.51 &75.57 &46.40 &79.60 &49.39 &70.32 &6.75\\
\midrule
 \textit{w/ QLoRA} & 512 & 4-16-16 &16-16-16& 66.59 &49.26 &75.20 &81.99 &76.06 &46.74 &79.49 &51.71 &72.27 &15.85\\
\noalign{\vspace{0.1em}}\hdashline[0.8pt/1pt]\noalign{\vspace{0.1em}}
\multirow{4}{*}{w/ GSQ-Tuning}  &\multirow{4}{*}{512} & 4-8-8 &8-8-8  & \default{\textbf{66.52}} &49.49 &74.92 &81.28 &75.89 &47.60 &79.49 &51.59 &71.90 &11.40\\
& & 4-7-7 &7-7-7  & \default{\textbf{66.33}} &48.89 &74.75 &81.41 &76.06 &47.00 &79.54 &51.74 &71.27 &9.95\\
& & 4-6-6 &6-6-6  & \default{\textbf{66.31}} &48.55 &75.51 &80.80 &76.42 &46.00 &79.60 &51.64 &71.98&9.19\\
  & & 4-5-5 &5-5-5 & \default{64.86} &47.44 &73.15 &76.85 &75.62 &47.00 &79.33 &49.18 &70.32 &8.25\\
\bottomrule
\end{tabular}}}
\end{table*}
\begin{table*}[!t]
\renewcommand\arraystretch{1.0}
\centering
\caption{$0$-shot commonsense QA accuracy (\%) across different bits and rank on llama2-13B.}
\label{tab:llama2-13b}
\setlength{\tabcolsep}{1.2mm}
{\resizebox{0.98\textwidth}{!}{
\begin{tabular}{lcccccccccccccc|c}
\noalign{\vspace{0.3em}}
\toprule
\noalign{\vspace{0.1em}}
\textbf{Method} & \textbf{rank}& LLMs branch & low-rank branch &\textbf{Avg.} & \textbf{ARC-c} & \textbf{ARC-e} & \textbf{BoolQ} & \textbf{HellaS.} & \textbf{OBQA} & \textbf{PIQA} & \textbf{SCIQ.} & \textbf{WinoG.} & \textbf{Mem. (G)} \\
\midrule
\noalign{\vspace{0.1em}}
 LLaMA2-13B    &-     & 16-16-None & w/o&66.65 &48.81 &76.47 &82.45 &79.67 &44.80 &80.36 &48.31 &72.38 &25.70\\
 \noalign{\vspace{0.1em}}\hdashline[0.8pt/1pt]\noalign{\vspace{0.1em}}
 \textit{w/ QLoRA} & 16 &4-16-16 & 16-16-16 & 67.32 &49.74 &76.98 &82.94 &78.85 &46.00 &80.52 &50.36 &73.16 &16.56\\
\noalign{\vspace{0.1em}}\hdashline[0.8pt/1pt]\noalign{\vspace{0.1em}}
\multirow{4}{*}{w/ GSQ-Tuning}  &\multirow{4}{*}{16} &4-8-8 & 8-8-8  & \default{\textbf{67.35}} &49.83 &77.06 &83.09 &78.89 &46.00 &80.47 &50.31 &73.16 &11.13\\
& & 4-7-7 & 7-7-7   & \default{\textbf{67.29}} &49.91 &76.94 &83.03 &78.90 &45.40 &80.58 &50.61 &73.01 &10.58\\
& & 4-6-6 & 6-6-6   & \default{\textbf{67.23}} &49.66 &76.98 &82.75 &78.79 &46.00 &80.47 &50.05 &73.16 &10.03\\
  & & 4-5-5 & 5-5-5  & \default{66.57} &49.57 &76.43 &81.62 &77.98 &45.40 &80.09 &49.39 &72.06 &9.47\\
\midrule
\noalign{\vspace{0.1em}}
 \textit{w/ QLoRA} & 32 &4-16-16 & 16-16-16 & 67.47 &49.83 &77.02 &83.24 &78.92 &46.20 &80.58 &50.77 &73.24 &16.85\\
\noalign{\vspace{0.1em}}\hdashline[0.8pt/1pt]\noalign{\vspace{0.1em}}
\multirow{4}{*}{w/ GSQ-Tuning}  &\multirow{4}{*}{32} & 4-8-8 & 8-8-8  & \default{\textbf{67.49}} &49.83 &76.98 &83.15 &78.94 &45.60 &80.79 &51.07 &73.56 &11.42\\
& & 4-7-7 & 7-7-7  & \default{\textbf{67.38}} &50.17 &77.06 &82.81 &78.99 &45.40 &80.79 &50.46 &73.40  &10.87\\
& & 4-6-6 & 6-6-6  & \default{\textbf{67.35}} &49.83 &77.06 &83.09 &78.89 &46.00 &80.47 &50.31 &73.16 &10.31\\
  & & 4-5-5 & 5-5-5 & \default{66.65} &48.38 &76.18 &82.08 &78.07 &45.60 &80.36 &49.74 &72.77 &9.76 \\
\midrule
 \textit{w/ QLoRA} & 64 &4-16-16 & 16-16-16 & 67.61 &49.66 &77.23 &83.30 &78.95 &45.40 &80.74 &51.59 &73.24 &17.42\\
\noalign{\vspace{0.1em}}\hdashline[0.8pt/1pt]\noalign{\vspace{0.1em}}
\multirow{4}{*}{w/ GSQ-Tuning}  &\multirow{4}{*}{64} & 4-8-8 & 8-8-8   & \default{\textbf{67.48}} &49.57 &77.40 &82.87 &78.88 &46.20 &80.90 &50.72 &73.32 &11.99\\
& & 4-7-7 & 7-7-7   & \default{\textbf{67.43}} &49.74 &77.27 &82.91 &78.89 &46.00 &80.90 &50.61 &73.09 &11.44\\
& & 4-6-6 & 6-6-6   & \default{\textbf{67.35}} &49.66 &77.27 &82.75 &78.66 &79.05 &80.90 &50.97 &73.16 &10.89\\
  & & 4-5-5 & 5-5-5 & \default{66.97} &49.91 &76.60 &81.87 &78.15 &46.20 &80.41 &49.54 &73.09 &10.33\\
\midrule
 \textit{w/ QLoRA} & 128&4-16-16 & 16-16-16& 67.61 &50.34 &77.40 &83.55 &78.89 &46.00 &80.85 &50.92 &72.93&18.56\\
\noalign{\vspace{0.1em}}\hdashline[0.8pt/1pt]\noalign{\vspace{0.1em}}
\multirow{4}{*}{w/ GSQ-Tuning}  &\multirow{4}{*}{128} & 4-8-8 & 8-8-8   & \default{\textbf{67.62}} &50.34 &77.06 &83.18 &78.96 &46.40 &80.69 &50.92 &73.40&13.14\\
& & 4-7-7 & 7-7-7   & \default{\textbf{67.57}} &50.43 &77.36 &83.06 &79.05 &45.60 &80.85 &51.28 &72.93 &12.58\\
& & 4-6-6 & 6-6-6   & \default{\textbf{67.53}} &50.43 &77.31 &83.15 &78.81 &45.80 &80.58 &50.97 &73.16 &12.03\\
  & & 4-5-5 & 5-5-5  & \default{67.10} &49.49 &76.81 &82.08 &78.22 &46.40 &80.03 &50.56 &73.24 &11.48\\
\midrule
 \textit{w/ QLoRA} & 256 &4-16-16 & 16-16-16 & 67.91 &50.77 &77.36 &83.64 &78.88 &46.60 &80.74 &51.69 &73.64&20.85\\
\noalign{\vspace{0.1em}}\hdashline[0.8pt/1pt]\noalign{\vspace{0.1em}}
\multirow{4}{*}{w/ GSQ-Tuning}  &\multirow{4}{*}{256} & 4-8-8 & 8-8-8   & \default{\textbf{67.84}} &51.11 &77.06 &83.82 &78.80 &46.40 &80.69 &52.00 &72.85 &15.42\\
& &4-7-7 & 7-7-7   & \default{\textbf{67.74}} &50.77 &77.31 &83.79 &78.84 &46.00 &80.63 &51.89 &72.69 &14.87\\
& & 4-6-6 & 6-6-6   & \default{\textbf{67.68}} &50.77 &77.19 &83.49 &78.82 &46.00 &80.58 &51.38 &73.24 &14.32\\
  & & 4-5-5 & 5-5-5  & \default{67.22} &50.85 &75.84 &82.11 &78.21 &46.00 &80.36 &50.92 &73.48 &13.76\\
\midrule
 \textit{w/ QLoRA} & 512 &4-16-16 & 16-16-16 & 67.94 &50.60 &77.48 &83.88 &79.00 &46.40 &80.74 &52.05 &73.40&25.43\\
\noalign{\vspace{0.1em}}\hdashline[0.8pt/1pt]\noalign{\vspace{0.1em}}
\multirow{4}{*}{w/ GSQ-Tuning}  &\multirow{4}{*}{512} & 4-8-8 & 8-8-8   & \default{\textbf{67.92}} &51.02 &77.27 &83.27 &79.04 &46.40 &81.01 &51.79 &73.56 &20.00\\
& & 4-7-7 & 7-7-7   & \default{\textbf{67.90}} &51.19 &77.15 &83.79 &78.82 &46.80 &80.69 &51.79 &73.01 &19.45\\
& & 4-6-6 & 6-6-6   & \default{\textbf{67.82}} &51.02 &77.02 &83.85 &78.93 &46.20 &80.90 &51.54 &73.09 &18.89\\
  & & 4-5-5 & 5-5-5  & \default{67.39} &50.94 &76.68 &82.29 &78.39 &46.20 &80.41 &51.69 &72.53 &18.34\\
\bottomrule
\end{tabular}}}
\end{table*}
\begin{table*}[!t]
\renewcommand\arraystretch{1.0}
\centering
\caption{$0$-shot commonsense QA accuracy (\%) across different bits and rank on llama2-70B.}
\label{tab:llama2-70b}
\setlength{\tabcolsep}{1.2mm}
{\resizebox{0.98\textwidth}{!}{
\begin{tabular}{lcccccccccccccc|c}
\noalign{\vspace{0.3em}}
\toprule
\noalign{\vspace{0.1em}}
\textbf{Method} & \textbf{rank}& LLMs branch & low-rank branch &\textbf{Avg.} & \textbf{ARC-c} & \textbf{ARC-e} & \textbf{BoolQ} & \textbf{HellaS.} & \textbf{OBQA} & \textbf{PIQA} & \textbf{SCIQ.} & \textbf{WinoG.} & \textbf{Mem. (G)} \\
\midrule
\noalign{\vspace{0.1em}}
 LLaMA2-70B    &-     &  16-16-None &w/o & 70.68 &56.91 &80.05 &85.78 &83.59 &48.60 &82.48 &48.67 &79.40 &137.42\\
 \noalign{\vspace{0.1em}}\hdashline[0.8pt/1pt]\noalign{\vspace{0.1em}}
 \textit{w/ QLoRA} & 16 & 4-16-16 &16-16-16 & 71.72 &58.62 &81.44 &86.39 &83.92 &49.80 &83.03 &50.46 &80.11 &63.90\\
\noalign{\vspace{0.1em}}\hdashline[0.8pt/1pt]\noalign{\vspace{0.1em}}
\multirow{4}{*}{w/ GSQ-Tuning}  &\multirow{4}{*}{16} & 4-8-8&8-8-8  & \default{\textbf{71.65}} &58.62 &81.23 &86.36 &83.87 &49.60 &83.19 &50.41 &79.95 &49.17\\
& & 4-7-7&7-7-7  & \default{\textbf{71.63}} &58.87 &81.57 &86.24 &83.89 &49.20 &83.19 &50.46 &79.64 &47.44\\
& & 4-6-6&6-6-6  & \default{\textbf{71.58}} &58.62 &81.36 &86.15 &83.84 &49.60 &82.97 &50.41 &79.64 &45.72\\
  & & 4-5-5&5-5-5 & \default{71.02} &57.34 &80.56 &85.93 &83.75 &49.00 &82.59 &49.33 &79.64 &43.99\\
\midrule
\noalign{\vspace{0.1em}}
 \textit{w/ QLoRA} & 32 &4-16-16 &16-16-16 & 71.84 &59.13 &81.82 &86.27 &83.88 &49.20 &83.03 &51.02 &80.35 &64.87\\
\noalign{\vspace{0.1em}}\hdashline[0.8pt/1pt]\noalign{\vspace{0.1em}}
\multirow{4}{*}{w/ GSQ-Tuning}  &\multirow{4}{*}{32} & 4-8-8&8-8-8  & \default{\textbf{71.78}} &59.04 &81.90 &86.33 &83.89 &49.00 &83.19 &51.07 &79.79 &50.17\\
& & 4-7-7&7-7-7  & \default{\textbf{71.76}} &59.30 &81.61 &86.18 &83.98 &49.00 &83.19 &51.02 &79.79 &48.44\\
& & 4-6-6&6-6-6  & \default{\textbf{71.60}} &58.96 &81.36 &86.15 &83.87 &48.80 &83.03 &51.02 &79.64 &46.72\\
  & & 4-5-5&5-5-5 & \default{71.26} &57.59 &80.85 &86.15 &83.93 &49.00 &83.13 &50.00 &79.40 &44.99\\
\midrule
 \textit{w/ QLoRA} & 64 & 4-16-16 &16-16-16 & 72.22 &59.81 &82.20 &86.51 &83.89 &50.40 &83.13 &51.48 &80.35 &66.82\\
\noalign{\vspace{0.1em}}\hdashline[0.8pt/1pt]\noalign{\vspace{0.1em}}
\multirow{4}{*}{w/ GSQ-Tuning}  &\multirow{4}{*}{64} & 4-8-8&8-8-8  & \default{\textbf{72.20}} &59.90 &82.32 &86.51 &83.90 &50.20 &83.08 &51.59 &80.11 &52.17\\
& & 4-7-7&7-7-7  & \default{\textbf{72.18}} &59.81 &82.28 &86.39 &83.88 &50.20 &83.13 &51.54 &80.19 &50.44\\
& & 4-6-6&6-6-6  & \default{\textbf{72.10}}  &59.39 &82.15 &86.51 &83.94 &50.00 &83.30 &50.92 &80.58 &48.71\\
  & & 4-5-5&5-5-5 & \default{71.70} &58.87 &81.48 &85.90 &83.91 &49.60 &82.81 &50.67 &80.43 &46.98\\
\midrule
 \textit{w/ QLoRA} & 128 & 4-16-16 &16-16-16& 72.39& 60.67 &82.37 &86.88 &84.05 &49.20 &83.19 &52.15 &80.66&70.96\\
\noalign{\vspace{0.1em}}\hdashline[0.8pt/1pt]\noalign{\vspace{0.1em}}
\multirow{4}{*}{w/ GSQ-Tuning}  &\multirow{4}{*}{128} & 4-8-8&8-8-8  &\default{\textbf{72.37}} &60.75 &82.49 &87.00 &83.94 &49.40 &83.08 &52.15 &80.19 &56.16\\
& & 4-7-7&7-7-7  & \default{\textbf{72.32}} &60.41 &82.45 &86.94 &83.94 &49.00 &83.08 &52.15 & 80.58 &54.43\\
& & 4-6-6&6-6-6  & \default{\textbf{72.28}}  &59.81 &82.45 &86.91 &83.99 &49.60 &83.35 &51.89 & 80.27 &52.70\\
  & & 4-5-5&5-5-5 & \default{71.85} &59.47 &81.90 &86.48 &83.82 &48.20 &83.08 &51.02 & 80.82&50.97\\
\bottomrule
\end{tabular}}}
\end{table*}
\begin{table*}[!t]
\renewcommand\arraystretch{1.0}
\centering
\caption{$0$-shot commonsense QA accuracy (\%) across different bits and rank on llama3-3B.}
\label{tab:llama3-3b}
\setlength{\tabcolsep}{1.2mm}
{\resizebox{0.98\textwidth}{!}{
\begin{tabular}{lcccccccccccccc|c}
\noalign{\vspace{0.3em}}
\toprule
\noalign{\vspace{0.1em}}
\textbf{Method} & \textbf{rank}& LLMs branch &low-rank branch &\textbf{Avg.} & \textbf{ARC-c} & \textbf{ARC-e} & \textbf{BoolQ} & \textbf{HellaS.} & \textbf{OBQA} & \textbf{PIQA} & \textbf{SCIQ.} & \textbf{WinoG.} & \textbf{Mem. (G)} \\
\midrule
\noalign{\vspace{0.1em}}
 LLaMA3-3B    &-     &  16-16-None & w/o & 64.13 &46.25 &74.62 &77.68 &76.01 &44.20 &79.11 &46.11 &69.06 &6.42\\
 \noalign{\vspace{0.1em}}\hdashline[0.8pt/1pt]\noalign{\vspace{0.1em}}
 \textit{w/ QLoRA} & 16 & 4-16-16 & 16-16-16 & 65.05 &47.53 &75.17 &78.59 &76.09 &44.00 &79.54 &49.44 &70.09 &6.42\\
\noalign{\vspace{0.1em}}\hdashline[0.8pt/1pt]\noalign{\vspace{0.1em}}
\multirow{4}{*}{w/ GSQ-Tuning}  &\multirow{4}{*}{16} & 4-8-8 & 8-8-8  & \default{\textbf{65.10}}& 47.53 &74.71 &78.35 &75.99 &45.00 &79.65 &49.28 &70.32 &3.57\\
& & 4-7-7 & 7-7-7  & \default{\textbf{64.96}} &47.18 &75.21 &78.10 &75.98 &44.80 &79.27 &49.95 &69.38 &3.34\\
& & 4-6-6 & 6-6-6  & \default{\textbf{64.87}} &46.84 &73.78 &78.07 &75.88 &45.80 &79.22 &49.39 &70.01 &3.11\\
  & & 4-5-5 & 5-5-5 & \default{63.97} &46.76 &72.64 &75.78 &74.95 &45.20 &79.05 &48.62 &68.75 &2.88\\
\midrule
\noalign{\vspace{0.1em}}
 \textit{w/ QLoRA} & 32 &4-16-16 & 16-16-16 & 65.24 &47.27 &75.04 &78.87 &76.11 &44.60 &79.76 &49.95 &70.32 &6.54\\
\noalign{\vspace{0.1em}}\hdashline[0.8pt/1pt]\noalign{\vspace{0.1em}}
\multirow{4}{*}{w/ GSQ-Tuning}  &\multirow{4}{*}{32} & 4-8-8 & 8-8-8 & \default{\textbf{65.45}} &48.12 &74.71 &78.38 &76.14 &46.00 &79.71 &49.64 &70.96 &3.69\\
& & 4-7-7 & 7-7-7  & \default{\textbf{65.43}} &47.35 &74.20 &78.99 &75.84 &46.00 &79.92 &49.59 &71.59 &3.46\\
& & 4-6-6 & 6-6-6 & \default{\textbf{65.01}} &47.44 &74.62 &78.65 &76.03 &44.00 &79.60 &50.05 &69.69 &3.23\\
  & & 4-5-5 & 5-5-5 & \default{64.00} &44.97 &73.32 &75.29 &74.95 &44.60 &79.27 &48.93 &70.24 &3.00\\
\midrule
 \textit{w/ QLoRA} & 64 & 4-16-16 & 16-16-16 & 65.69 &47.14 &74.75 &79.50 &76.46 &45.50 &79.63 &50.26 &71.32 &6.78\\
\noalign{\vspace{0.1em}}\hdashline[0.8pt/1pt]\noalign{\vspace{0.1em}}
\multirow{4}{*}{w/ GSQ-Tuning}  &\multirow{4}{*}{64} & 4-8-8 & 8-8-8  & \default{\textbf{65.60}} &48.12 &74.24 &79.72 &76.00 &45.80 &79.60 &49.69 &71.67 &3.93\\
& & 4-7-7 & 7-7-7  & \default{\textbf{65.47}} &47.78 &74.71 &79.51 &76.09 &45.80 &79.60 &49.80 &70.48 &3.70\\
& & 4-6-6 & 6-6-6  & \default{\textbf{65.39}} &47.70 &74.58 &79.24 &76.05 &44.60 &79.60 &50.41 &70.96 &3.47\\
  & &4-5-5 & 5-5-5 & \default{64.18} &45.14 &72.69 &75.20 &75.27 &46.40 &79.65 &48.62 &70.48 &3.24\\
\midrule
 \textit{w/ QLoRA} & 128 & 4-16-16 & 16-16-16& 65.84 &48.24 &74.91 &79.78 &76.27 &45.52 &79.77 &50.48 &71.79 &6.76\\
\noalign{\vspace{0.1em}}\hdashline[0.8pt/1pt]\noalign{\vspace{0.1em}}
\multirow{4}{*}{w/ GSQ-Tuning}  &\multirow{4}{*}{128} & 4-8-8 & 8-8-8  & \default{\textbf{65.79}} &48.12 &74.83 &80.28 &75.96 &45.80 &79.54 &50.61 &71.19 &4.41\\
& & 4-7-7 & 7-7-7  & \default{\textbf{65.69}} &48.04 &74.87 &79.79 &76.08 &45.00 &79.49 &50.61 &71.67 &4.18\\
& & 4-6-6 & 6-6-6  & \default{\textbf{65.58}} &47.87 &74.54 &80.09 &76.05 &45.40 &79.38 &50.10 &71.27 &3.95\\
  & & 4-5-5 & 5-5-5 & \default{64.46} &46.50 &72.77 &75.99 &75.31 &46.60 &79.00 &48.98 &70.56 &3.72\\
\midrule
 \textit{w/ QLoRA} & 256 &4-16-16 & 16-16-16 & 66.12 &48.33 &75.00 &80.94 &76.37 &45.61 &79.97 &51.13 &71.64 &7.61\\
\noalign{\vspace{0.1em}}\hdashline[0.8pt/1pt]\noalign{\vspace{0.1em}}
\multirow{4}{*}{w/ GSQ-Tuning}  &\multirow{4}{*}{256} & 4-8-8 & 8-8-8  & \default{\textbf{66.19}} &48.55 &75.13 &80.76 &76.14 &47.00 &79.38 &50.72 &71.82 &5.37\\
& & 4-7-7 & 7-7-7  & \default{\textbf{65.96}} &48.46 &75.08 &80.43 &76.04 &45.60 &79.76 &50.72 &71.59 &5.13\\
& & 4-6-6 & 6-6-6  & \default{\textbf{65.90}} &48.38 &74.16 &79.94 &75.81 &46.80 &79.43 &50.87 &71.82 &4.90\\
  & & 4-5-5 & 5-5-5 & \default{64.59} &46.33 &72.60 &76.51 &75.57 &46.40 &79.60 &49.39 &70.32 &4.67\\
\midrule
 \textit{w/ QLoRA} & 512 & 4-16-16 & 16-16-16 & 66.59 &49.26 &75.20 &81.99 &76.06 &46.74 &79.49 &51.71 &72.27&9.73\\
\noalign{\vspace{0.1em}}\hdashline[0.8pt/1pt]\noalign{\vspace{0.1em}}
\multirow{4}{*}{w/ GSQ-Tuning}  &\multirow{4}{*}{512} & 4-8-8 & 8-8-8  & \default{\textbf{66.52}} &49.49 &74.92 &81.28 &75.89 &47.60 &79.49 &51.59 &71.90&7.28\\
& & 4-7-7 & 7-7-7  & \default{\textbf{66.33}} &48.89 &74.75 &81.41 &76.06 &47.00 &79.54 &51.74 &71.27&7.05\\
& & 4-6-6 & 6-6-6 & \default{\textbf{66.31}} &48.55 &75.51 &80.80 &76.42 &46.00 &79.60 &51.64 &71.98 &6.82\\
  & & 4-5-5 & 5-5-5 & \default{64.86} &47.44 &73.15 &76.85 &75.62 &47.00 &79.33 &49.18 &70.32 &6.59\\
\bottomrule
\end{tabular}}}
\end{table*}
\begin{table*}[!t]
\renewcommand\arraystretch{1.0}
\centering
\caption{$0$-shot commonsense QA accuracy (\%) across different bits and rank on llama3-8B.}
\label{tab:llama3-8b}
\setlength{\tabcolsep}{1.2mm}
{\resizebox{0.98\textwidth}{!}{
\begin{tabular}{lcccccccccccccc|c}
\noalign{\vspace{0.3em}}
\toprule
\noalign{\vspace{0.1em}}
\textbf{Method} & \textbf{rank}& LLMs branch &low-rank branch &\textbf{Avg.} & \textbf{ARC-c} & \textbf{ARC-e} & \textbf{BoolQ} & \textbf{HellaS.} & \textbf{OBQA} & \textbf{PIQA} & \textbf{SCIQ.} & \textbf{WinoG.} & \textbf{Mem. (G)} \\
\midrule
\noalign{\vspace{0.1em}}
 LLaMA3-8B    &-     &  16-16-None & w/o & 67.18 &53.50 &77.74 &81.13 &79.20 &45.00 &80.63 &47.03 &73.24 &15.01\\
 \noalign{\vspace{0.1em}}\hdashline[0.8pt/1pt]\noalign{\vspace{0.1em}}
 \textit{w/ QLoRA} & 16 & 4-16-16 &16-16-16 & 68.14 &54.52 &79.50 &83.43 &78.66 &44.80 &80.85 &50.00 &73.32 &10.71\\
\noalign{\vspace{0.1em}}\hdashline[0.8pt/1pt]\noalign{\vspace{0.1em}}
\multirow{4}{*}{w/ GSQ-Tuning}  &\multirow{4}{*}{16} &  4-8-8 & 8-8-8  & \default{\textbf{68.16}} &54.61 &79.84 &83.70 &78.58 &44.80 &80.79 &49.85 &73.16 &7.03\\
& & 4-7-7 & 7-7-7  & \default{\textbf{68.00}} &54.01 &79.29 &83.46 &78.65 &45.00 &80.85 &49.80 &73.01 &6.65\\
& & 4-6-6 & 6-6-6  & \default{\textbf{67.74}} &54.01 &78.70 &83.09 &78.49 &44.00 &80.90 &49.44 &73.32 &6.26\\
  & & 4-5-5 & 5-5-5 & \default{66.51} &51.54 &77.27 &81.99 &77.00 &44.40 &78.84 &48.46 &72.61 &5.87\\
\midrule
\noalign{\vspace{0.1em}}
 \textit{w/ QLoRA} & 32 &4-16-16 &16-16-16 & 68.31 &55.55 &80.39 &83.36 &78.65 &44.60 &81.28 &50.05 &72.61 &11.02\\
\noalign{\vspace{0.1em}}\hdashline[0.8pt/1pt]\noalign{\vspace{0.1em}}
\multirow{4}{*}{w/ GSQ-Tuning}  &\multirow{4}{*}{32} & 4-8-8 & 8-8-8  & \default{\textbf{68.45}} &55.72 &80.22 &83.43 &78.60 &45.00  &81.18 &50.20 &73.32 &7.23\\
& & 4-7-7 & 7-7-7  & \default{\textbf{68.29}} &54.95 &80.13 &83.36 &78.53 &44.80 &81.01 &50.20 &73.32 &6.84\\
& & 4-6-6 & 6-6-6  & \default{\textbf{68.08}} &55.29 &79.29 &83.55 &78.28 &45.80 &81.07 &49.39 &71.98 &6.46\\
  & & 4-5-5 & 5-5-5 & \default{66.48} &51.71 &77.69 &82.11 &76.91 &44.20 &79.43 &48.16 &71.67 &6.07\\
\midrule
 \textit{w/ QLoRA} & 64 & 4-16-16 &16-16-16 & 68.45 &55.63 &80.13 &83.67 &78.78 &44.80 &81.28 &50.41 &72.93 &11.64\\
\noalign{\vspace{0.1em}}\hdashline[0.8pt/1pt]\noalign{\vspace{0.1em}}
\multirow{4}{*}{w/ GSQ-Tuning}  &\multirow{4}{*}{64} & 4-8-8 & 8-8-8  & \default{\textbf{68.61}} &55.97 &80.22 &83.61 &78.68 &45.20 &81.50 &50.41 &73.32 &7.63\\
& & 4-7-7 & 7-7-7  & \default{\textbf{68.57}}  &55.97 &80.68 &83.73 &78.84 &45.20 &81.01 &50.26 &72.85 &7.24\\
& & 4-6-6 & 6-6-6  & \default{\textbf{68.22}}  &55.55 &79.29 &83.67 &78.47 &44.80 &80.90 &50.05 &73.09 &6.86\\
  & & 4-5-5 & 5-5-5 & \default{66.69} &54.10 &77.99 &81.65 &77.12 &43.80 &79.54 &47.90 &71.43 &6.47\\
\midrule
 \textit{w/ QLoRA} & 128 & 4-16-16 &16-16-16 & 68.77 &56.14 &80.56 &83.98 &79.03 &45.60 &81.34 &50.56 &72.93 &12.13\\
\noalign{\vspace{0.1em}}\hdashline[0.8pt/1pt]\noalign{\vspace{0.1em}}
\multirow{4}{*}{w/ GSQ-Tuning}  &\multirow{4}{*}{128} & 4-8-8 & 8-8-8  & \default{\textbf{68.72}} &56.57 &80.22 &83.82 &78.80 &45.40 &81.23 &50.41 &73.32 &8.43\\
& & 4-7-7 & 7-7-7  & \default{\textbf{68.71}} &56.48 &80.18 &83.88 &78.78 &45.80 &81.34 &50.36 &72.93 &8.04\\
& & 4-6-6 & 6-6-6  & \default{\textbf{68.67}} &56.91 &79.50 &83.79 &78.71 &46.60 &80.52 &50.36 &73.01 &7.66\\
  & &4-5-5 & 5-5-5 & \default{66.92} &52.47 &78.45 &82.63 &77.22 &44.60 &79.49 &48.52 &71.98&7.27\\
\midrule
 \textit{w/ QLoRA} & 256 & 4-16-16 &16-16-16 & 69.09 &56.74 &80.35 &84.56 &79.02 &45.20 &81.83 &50.92 &74.11&13.81\\
\noalign{\vspace{0.1em}}\hdashline[0.8pt/1pt]\noalign{\vspace{0.1em}}
\multirow{4}{*}{w/ GSQ-Tuning}  &\multirow{4}{*}{256} & 4-8-8 & 8-8-8  & \default{\textbf{69.04}} &56.57 &80.85 &84.07 &78.97 &45.40 &81.45 &51.28 &73.72 &10.03 \\
& & 4-7-7 & 7-7-7  & \default{\textbf{69.00}} &56.83 &80.89 &84.25 &78.96 &45.60 &81.50 &50.46 &73.56 &9.64\\
& & 4-6-6 & 6-6-6  & \default{\textbf{68.84}} &56.74 &79.80 &83.98 &78.84 &46.40 &81.12 &50.77 &73.09 &9.26\\
  & & 4-5-5 & 5-5-5 & \default{67.54} &53.33 &78.49 &83.21 &77.38 &44.60 &79.98 &48.93 &73.64 &8.87\\
\midrule
 \textit{w/ QLoRA} & 512 &4-16-16 &16-16-16 & 69.18 &57.17 &80.30 &84.65 &79.28 &46.40 &81.07 &50.36 &74.27 &16.81\\
\noalign{\vspace{0.1em}}\hdashline[0.8pt/1pt]\noalign{\vspace{0.1em}}
\multirow{4}{*}{w/ GSQ-Tuning}  &\multirow{4}{*}{512} & 4-8-8 & 8-8-8  & \default{\textbf{69.24}} &56.48 &80.47 &85.35 &79.13 &45.40 &81.56 &51.54 &74.03 &13.23\\
& & 4-7-7 & 7-7-7 & \default{\textbf{69.16}} &56.40 &80.68 &85.26 &79.10 &45.80 &81.28 &51.13 &73.64 &12.84\\
& & 4-6-6 & 6-6-6  & \default{\textbf{69.01}} &56.57 &80.01 &84.56 &78.84 &45.80 &81.23 &51.64 &73.48 &12.45\\
  & & 4-5-5 & 5-5-5 & \default{67.90} &54.38 &78.60 &83.97 &78.08 &45.30 &80.24 &50.00 &72.76 &12.07\\
\bottomrule
\end{tabular}}}
\end{table*}
\subsection{Detailed results on different rank setting:}
\label{sec:detailed_results}
Here, we also report the results of our GSQ-Tuning on different LlaMA model, including LlaMA2-7B (Tab.\ref{tab:llama2-7b}), LlaMA2-13B (Tab.\ref{tab:llama2-13b}), LlaMA2-70B(Tab.\ref{tab:llama2-70b}), LlaMA3-3B(Tab.\ref{tab:llama3-3b}), and LlaMA3-8B(Tab.\ref{tab:llama3-8b}). The results consistently demonstrated the effectiveness and efficiency of GSQ-Tuning.

\subsection{Comparison with FP8 with 64 rank}
Here, we compare the designed GSE data format with FP8 in fully quantized fine-tuning framework with 32 rank setting. As shown in Tab.~\ref{tab:copare_fp8_r64}, the results still demonstrate that the designed GSE implemented in our GSQ-Tuning method achieves superior fine-tuning performance compared to FP8 while significantly reducing computation efficiency. Even under 5-bit settings, GSQ-Tuning maintains fine-tuning performance on par with FP8, validating its effectiveness.
\begin{table*}[!t]
\renewcommand\arraystretch{1.0}
\centering
\caption{$0$-shot accuracy comparison with FP8 in different quantization bits in 64 rank setting.}
\label{tab:copare_fp8_r64}
\setlength{\tabcolsep}{1.2mm}
{\resizebox{0.98\textwidth}{!}{
\begin{tabular}{lccccccccccccc|c}
\noalign{\vspace{0.3em}}
\toprule
\noalign{\vspace{0.1em}}
\textbf{Method} & LLMs branch &low-rank branch  &\textbf{Avg.} & \textbf{ARC-c} & \textbf{ARC-e} & \textbf{BoolQ} & \textbf{HellaS.} & \textbf{OBQA} & \textbf{PIQA} & \textbf{SCIQ.} & \textbf{WinoG.} & \textbf{Mem. (G)} \\
\midrule
\noalign{\vspace{0.1em}}
 LLaMA2-7B         &  16-16-None & w/o & 64.13 &46.25 &74.62 &77.68 &76.01 &44.20 &79.11 &46.11 &69.06 &13.20\\
 \textit{w/ QLoRA} &  4-16-16 & 16-16-16 & 65.69 &47.14 &74.75 &79.50 &76.46 &45.50 &79.63 &50.26 &71.32 &9.73\\
\noalign{\vspace{0.1em}}\hdashline[0.8pt/1pt]\noalign{\vspace{0.1em}}
w/ FP8 & 4-8-8 & 8-8-8 & 64.46 &46.84 &73.61 &77.83 &76.03 &44.60 &79.65 &47.80 &69.38 &6.88\\
\noalign{\vspace{0.1em}}\hdashline[0.8pt/1pt]\noalign{\vspace{0.1em}}
\multirow{3}{*}{w/ GSQ-Tuning}  &4-8-8 & 8-8-8  & 65.60 &48.12 &74.24 &79.72 &76.00 &45.80 &79.60 &49.69 &71.67 &6.88\\
& 4-6-6 & 6-6-6  & 65.39 &47.70 &74.58 &79.24 &76.05 &44.60 &79.60 &50.41 &70.96 &6.17\\
  & 4-5-5 & 5-5-5 & 64.18 &45.14 &72.69 &75.20 &75.27 &46.40 &79.65 &48.62 &70.48 &5.81\\
\midrule
 LLaMA3-8B         & 16-16-None & w/o & 67.18 &53.50 &77.74 &81.13 &79.20 &45.00 &80.63 &47.03 &73.24 &15.01\\
 \textit{w/ QLoRA} & 4-16-16 & 16-16-16 & 68.45 &55.63 &80.13 &83.67 &78.78 &44.80 &81.28 &50.41 &72.93 &11.71\\
\noalign{\vspace{0.1em}}\hdashline[0.8pt/1pt]\noalign{\vspace{0.1em}}
w/ FP8  & 4-8-8 & 8-8-8  & 66.46 &50.77 &76.39 &81.38 &78.19 &43.40 &79.92 &47.29 &74.35 &7.63\\
\noalign{\vspace{0.1em}}\hdashline[0.8pt/1pt]\noalign{\vspace{0.1em}}
 \multirow{3}{*}{w/ GSQ-Tuning} & 4-8-8 & 8-8-8  & 68.61 &55.97 &80.22 &83.61 &78.68 &45.20 &81.50 &50.41 &73.32 &7.63\\
 & 4-6-6 & 6-6-6 & 68.22 &55.55 &79.29 &83.67 &78.47 &44.80 &80.90 &50.05 &73.09 &6.86\\
& 4-5-5 & 5-5-5 & 66.69 &54.10 &77.99 &81.65 &77.12 &43.80 &79.54 &47.90 &71.43 &6.47\\
\bottomrule
\end{tabular}}}
\end{table*}

\end{document}